\documentclass[10pt]{article}

\usepackage{graphicx} 
\usepackage[noend]{algpseudocode}
\usepackage{amsmath,amsthm}
\usepackage[linesnumbered,ruled,vlined]{algorithm2e}
\usepackage{todonotes}
\usepackage{enumitem}
\usepackage{subcaption} 
\usepackage{booktabs} 
\usepackage[ruled]{algorithm2e} 
\usepackage{amsfonts}
\usepackage[switch]{lineno} 
\usepackage{url}

\usepackage{authblk}

\title{Forming Diverse Teams from Sequentially Arriving People}

\author[1]{Faez Ahmed}
\affil[1]{Dept. of Mechanical Engineering\\
	University of Maryland\\
	College Park, Maryland 20742\\
    Email: faez00@umd.edu
    }	
 
\author[2]{John Dickerson}
 \affil[2]{Dept. of Computer Science\\
	University of Maryland\\
	College Park, Maryland 20742\\
    Email: john@cs.umd.edu
    }	
\author[3]{Mark Fuge}
\affil[3]{Dept. of Mechanical Engineering\\
	University of Maryland\\
	College Park, Maryland 20742\\
    Email: fuge@umd.edu
}

\date{}

\newcommand\cscreen[1]{c^{\textsc{S}}_{#1}}
\newcommand\cbonus[2]{c^{\textsc{B}}_{{#1},{#2}}}
\newcommand\OPT{\mathit{OPT}}
\newcommand\EntropyGain{\textsc{EG}}
\newcommand\PoDnum{\textsc{PoD}_{\#}}
\newcommand\PoDutil{\textsc{PoD}_{u}}

\newcommand{\eg}{{\em e.g.}}
\newcommand{\ie}{{\em i.e.}}
\newcommand{\etc}{{\em etc.}}
\newcommand{\etal}{{\em et~al.}}

\usepackage{amssymb}
\usepackage{fancyhdr}

\begin{document}
\maketitle    


\begin{abstract}
Collaborative work often benefits from having teams or organizations with heterogeneous members. In this paper, we present a method to form such diverse teams from people arriving sequentially over time. We define a monotone submodular objective function that combines the diversity and quality of a team and propose an algorithm to maximize the objective while satisfying multiple constraints. This allows us to balance both how diverse the team is and how well it can perform the task at hand. Using crowd experiments, we show that, in practice, the algorithm leads to large gains in team diversity. Using simulations, we show how to quantify the additional cost of forming diverse teams and how to address the problem of simultaneously maximizing diversity for several attributes (\eg{} country of origin, gender). Our method has applications in collaborative work ranging from team formation, the assignment of workers to teams in crowdsourcing, and reviewer allocation to journal papers arriving sequentially. Our code
is publicly accessible for further research.

\end{abstract}


\section{Introduction}\label{sec:intro}

\begin{figure}[t]
\centering
\includegraphics[height=3.6cm]{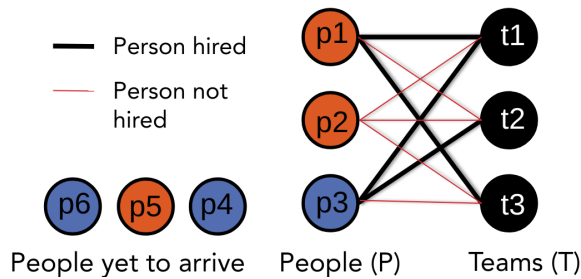}
 \caption{Bipartite graph of people arriving sequentially and teams requiring two workers each. People belong to two groups here (orange for p1, p2, p5 and blue for p3, p4, p6). Team $t1$ is matched to two people from different groups while team $t2$ and $t3$ are so far matched only to one person each. }

\label{fig:block}
\end{figure}

Collaborative work often benefits from having teams or organizations with diverse backgrounds and experiences~\cite{stirling2007general}. 
For example, studies have suggested that there is a positive relationship between diversity in a firm's knowledge base and its capability to innovate~\cite{hewlett2013diversity}. A large-scale study by Mckinsey~\cite{hunt2015diversity} looked at the relationship between the level of diversity (defined as a greater share of women and a more mixed ethnic/racial composition in the leadership of large companies) and company financial performance. They found that the companies in the top quartile of gender diversity were 15 percent more likely to have financial returns that were above their national industry median. Companies in the top quartile of racial/ethnic diversity were 30 percent more likely to have financial returns above their national industry median.
Firms or teams with employee diversity are often considered to be more competitive since such teams make the firm more open towards new ideas~\cite{paulus2016cultural}\textemdash for example, by increasing a firm's knowledge base and interaction between different competencies. As the cultural, educational and ethnic backgrounds among employees become more diverse, so does the knowledge base of the firm.

However, forming and maintaining diverse and high-quality teams over time can be challenging, in large part because people (whether in traditional firms or online collaborative groups) join and leave the firm sequentially, over time, rather than as one large cohort or pool. In contrast, if we knew ahead of time exactly when and who would be available to join different teams, then the problem reduces to the easier mathematical problem of static \textit{bipartite matching}: that is assigning a set of resources (people, in this case) to a set of tasks/groups (teams, in this case). If different people were better suited for some teams or tasks over others (say, they had a certain skill that was highly valued for a given team's task), then this is called \textit{weighted} bipartite matching, such that we assign people to teams such that the assignment maximizes the overall weight (or quality) of the matching. In practice, people can often be assigned to multiple teams or collaborative projects at the same time, up to some upper and lower limits (say, maximum of $b$ number of teams per person), which is referred to as \textit{weighted b-matching}. The widely-studied weighted b-matching problem occurs in all cases where a finite set of resources (\eg, people, computers, vehicles) needs to be matched to another finite set of resources (\eg, teams, tasks, trips) like team-formation and scientific peer-review (assigning people to review papers).

For collaborative work, we must handle two additional constraints not considered together by past matching approaches: 

\begin{enumerate}[noitemsep]
\item We do not know ahead of time exactly which future people will be available and need to decide at the moment whether to assign a newly arrived person to a team\textemdash \ie, we must match people to teams in \textit{real-time} rather than waiting to collect a pool of people and then matching everyone in that pool to teams in an \textit{offline} fashion; and
\item We want to encourage matching a \textit{diverse} subset of people to teams\textemdash \eg, teams where people are not only well-matched to the task but also have complementary expertise or relevant but different viewpoints.
\end{enumerate}
We refer to this as \textit{real-time, diverse, weighted b-matching}. Figure~\ref{fig:block} shows an illustration of this problem with three teams and people belonging to two groups. This setting is particularly important in practical implementations of collaborative work, where teams of people are formed to solve problems together.
Without real-time matching, for example, if one waits to match people to teams offline (\eg, by collecting a pool of people before assigning teams), then team starvation or worker retention issues arise where teams may sit dormant without progress or workers may leave while you wait to assign people to the teams. 
The problem of real-time diverse matching arises in many disciplines and problems, including: matching workers to firms~\cite{Horton17:Effects}, children to schools~\cite{Kurata15:Controlled,Drummond15:SAT}, reviewers to manuscripts~\cite{Charlin13:Toronto,Liu14:Robust}, donor organs to patients~\cite{Bertsimas13:Fairness,Dickerson15:FutureMatch} and residents to public housing~\cite{Benabbou18:Diversity}.
Specifically, this paper contributes to the following:
\begin{enumerate}[noitemsep]
\item We show how to formulate the real-time diverse bipartite $b$-matching problem as an optimization problem and demonstrate how our general formulation resolves, as a special case, to real-time person-to-team matching.
\item We present a simple approximation algorithm for performing real-time diverse matching. 
\item We demonstrate that the empirical performance of our simple, greedy allocation not only satisfies theoretical results but is often surprisingly close to optimal, in practice, on a variety of tasks including simulated test cases with known optima, and via Amazon Mechanical Turk experiments. 
\end{enumerate}

To enable practitioners to deploy this method for their own domain, we have provided the source code\footnote{\url{https://github.com/IDEALLab/onlinematching}} and encourage interested readers to use it.

\section{Related Work}\label{sec:rw}

Matching people to form diverse teams leverages the intersection of two past areas of research: the role of team diversity in collaborative work~(\S\ref{sec:rw-formation}) and how resource diversity is measured and used to form teams~(\S\ref{sec:rw-diversity}). In the context of this past work, this paper provides a practical, simple-to-implement, and high-performing method to perform diverse, real-time, b-matching that can enable diverse team formation when unknown people arrive sequentially over time.

\subsection{Diversity in teams} 
\label{sec:rw-formation}

Building effective teams is often defined as ``helping a work group become more effective in accomplishing its tasks and satisfying the needs of group members'' \cite{cummings2009organization}.
Prior research has explored what constitutes a successful team \cite{clutterbuck2011coaching}, how teams develop \cite{lenhardt2004coaching}, and how different selection criteria and competencies might lead a team to excel \cite{levi2015group}.
For example, effective teams may need diverse knowledge and skills\cite{holpp1999managing, humphrey2009developing, sassenberg2007some}, workers' attitudes, personalities \cite{barrick1998relating} and emotional intelligence \cite{jordan2002workgroup}. 
Team Diversity can include both task-related diversity (\eg, functional expertise, education, and organizational tenure) as well as bio-demographic diversity (\eg, age, gender, and race/ethnicity).
Task-related diversity has been reported to have a positive impact on team performance ~\cite{ross2010crowdworkers, ostergaard2011does, hunt2015diversity} although bio-demographic diversity is shown not to be significantly related to team performance~\cite{horwitz2007effects}. 

Non-diverse teams often emphasize on consensus-seeking behavior, which can result in suboptimal decision making, such as Groupthink~\cite{janis1971groupthink}. Team diversity can often circumvent this by bringing in differing perspectives and promoting healthy debates and dissents \cite{williams1998demography} with limited to no decrease in performance (\eg, \cite{harrison2002time}). For example, increased cognitive diversity can increase performance on complex and non-routine tasks~\cite{pelled1999exploring, cox1991managing}. In contrast, other researchers have argued for the benefits of homogeneous (non-diverse) teams which can include increased team cohesion and performance on certain tasks~\cite{bryne1966effect}.

In relation to that body of work, this paper provides an algorithm for organizations to control to what extent they wish to incorporate or emphasize various types of diversity when matching workers to teams.

\subsection{Measuring diversity and matching teams}\label{sec:rw-diversity}

While researchers have found benefits to encouraging team diversity (cognitive, task-based, \etc), one open question lies in how to rigorously and scalably \textit{form} teams (or, equivalently, match people to teams) to achieve that diversity. To do this, we first need to understand two areas of related research: (1) how is diversity measured and 2) how can one use those measures to form diverse teams?

Past researchers have measured diversity by defining some notion of \textit{coverage}\textemdash that is, a diverse set should covers the space of available variation. 
Mathematically, researchers have done so via the use of \textit{submodular functions}, which encode the notion of diminishing returns~\cite{Lin11:Class,lin2012learning}; that is, as one adds items to a set that are similar to previous items, one gains less utility if the existing items in the set already ``cover'' the characteristics added by that new item. For example, many previous diversity metrics used in the informational retrieval or search communities\textemdash including Maximum Marginal Relevance (MMR) \cite{carbonell1998use}, absorbing random walks \cite{zhu2007improving}, subtopic retrieval \cite{zhai2003beyond} and Determinantal Point Processes (DPP) \cite{kulesza2012determinantal}\textemdash are instances of submodular functions. These functions can model notions of coverage, representation, and diversity \cite{ahmed2018ranking} and they achieve the best results to date on common automatic document summarization benchmarks\textemdash \eg, at the Document Understanding Conference \cite{Lin11:Class,lin2012learning}.

Once one has an appropriate function for measuring diversity, one now has to use that function to form diverse teams. Wilde \etal{}~\cite{wilde2008teamology} proposed that diversity of a team can be measured by a count of the number of unique affinity groups present in the team. They provided a practical method to form teams based on the cognitive patterns of people in a personnel pool. However, their approach uses a diversity measure (which is similar to the Richness measure used in ecology) that does not account for affinity group variations within a team. Their heuristic approach does not simultaneously maximize quality and diversity, and cannot scale to cases with thousands of participants. While fully automated team formation algorithms have recently emerged to place people together in socially networked environments \cite{cruz2014group, anagnostopoulos2012online}, past approaches do not ensure or encourage diversity in any matchings, instead focusing only on how qualified the members are to the task (standard weighted b-matching) and meeting the cost/capacity constraint. However, in the offline case, Ahmed \etal{}~\cite{ijcai2017-6} provided an algorithm for \textit{diverse} b-matching applied to reviewer-paper matching of conference papers. Their matching occurred offline (where all people and tasks were known ahead of time) using a Mixed Integer Quadratic Program, rather than the real-time case that more realistically captures actual team formation in most firms or communities. They also proposed a pseudo-polynomial time algorithm, which guarantees to provide optimal solution for the offline matching problem using an auxiliary graph approach \cite{ahmadi2019algorithms}. In \cite{Cohen:2017:CDG:3068839.3068842}, authors study the offline diverse team formation problem and provide a polynomial method for approximating optimal team formation. They study a complementary definition of diversity, where the goal is to find teams that are close to a given distribution and not the team members being different from each other.

\subsection{Why form teams in real-time?}\label{sec:rw-online}
In contrast to offline team formation, real-time algorithms (also called online algorithms) are more appropriate for forming teams for tasks where a timeline exists with varying worker arrivals and departures. This paper contributes a means to form teams that are both diverse and formed in real-time. In real-time team assignment problems: (1) a firm has a fixed set of tasks/teams and a budget that specifies how many times the firm would like each task completed or how many teams it needs; (2)~new people arrive at the firm one at a time (in the case of regular hiring) and potentially the same person could arrive multiple times (\eg, in the case of freelancing or gig/shift work); and (3)~people must be assigned to a team immediately upon arrival (or rejected and not assigned to any team). The goal is to allocate people to teams in a way that maximizes the value of collaborative work all teams produce (\ie, solely maximizing utility).

But why do we need to form teams in real-time?
Doesn't one still need the team to be built to start performing its task? In such a case, the team members who arrive earlier need to wait for later team members to join the team. If they are waiting for other workers to join, why can they not just wait in a pool so that one can use an offline team formation algorithm? The answer to these questions relate to two main factors: (a) the type of the task, and (b) the compensation of the individuals. First, not all team tasks require the entire team to work synchronously. In tasks like conference paper reviewing, each team member works independently and then their output is aggregated. Real-time team formation works well for such tasks. However, even in tasks which require the team to work synchronously, real-time team formation can help when task timeliness and cost are constrained. To form the teams offline, one may have to create a large pool of workers and ask them all to wait until the pool is large enough. This means all the workers have to be paid while waiting and many may drop-out from the waiting room. In contrast, a real-time algorithm only requires that the selected workers wait, not the entire pool. This improvement in time comes at the cost of lower objective value, as offline matching will always be strictly better than the real-time matching method (assuming no drop-outs and ignoring the cost of waiting). Finally, in case of a batch of workers appearing at a time, our algorithm can be easily modified to use a submodular greedy method to rank order the entire batch and then use real-time matching algorithm outlined in Algorithm~\ref{algo:change}.

Few papers have studied the real-time or online task assignment problem. For instance, Roy \etal{}~\cite{basu2015task} proposed a framework for optimizing task assignment in knowledge-intensive crowdsourcing. They maximize overall task quality and minimize cost, with constraints on skill, cost, and tasks per worker. Unlike our work, they use an additive skill aggregation model~\cite{anagnostopoulos2012online} to calculate the total skill of a team of workers. 
The work closest in scope to our problem that of Schmitz \etal{}~\cite{schmitz2018online} who study the problem of both task assignment (finding which worker should do which task) and sequencing (identifying at what time each worker should contribute). Their model assumes that workers are available only at specific time slots and worker/task arrivals are not known apriori. In their work, the utility provided by each worker in a team or task is independent of other workers. 
This assumption can fail if previously arrived workers have similar skills and have already joined the team. In contrast to their method, we address a harder problem where every worker's utility depends on whoever else has already been accepted to the team. 

This paper addresses how to maximize \textit{both} utility and diversity\textemdash where we, similarly to past research, represent diversity using a submodular function. Mathematically, we essentially express the diverse real-time matching problem as a subset selection problem with multiple knapsack constraints.
Online matching and its generalization to set packing have been studied through the lens of theoretical computer science for nearly three decades~\cite{Karp90:Optimal}. These algorithms have been applied
to a multitude of tasks like online video summarization \cite{mirzasoleiman2018streaming}.
The algorithms we present in this paper draw motivation most heavily from recent work in online stochastic optimization with nonlinear objectives~\cite{Devanur12:Online,Agrawal14:Fast}, and from~\cite{7906050} in particular.

\section{Diversity in Matching}\label{sec:diversity}

This section introduces some of the more detailed mathematical notation needed to properly describe our algorithm for team formation in the next section. We flesh out in more precise detail how diversity is modeled and calculated via a submodular function and how this relates to matching people to teams.

We model the overall problem as maximizing a monotone submodular function over $b$-matchings in a bipartite graph $G ~=~ (P,~T,~E)$, where $P$ is a set of $M$ vertices (\eg, people) that arrive sequentially, $T$ is a set of $N$ vertices (e.g., teams) known a priori, and where no vertex $i$ (team or people) is incident to more than $b(i)$ edges in a proposed matching (\ie, we cannot assign a person $i$ to more than $b(i)$ teams at once), and $E$ is the set of edges between teams and people. Even the offline version of this problem is NP-Hard,so we focus on approximate submodular maximization and instead bound how close we can get to the optimal solution.
To incorporate diversity, we consider a scenario where left-side nodes (\eg, the people) are divided into $K$ groups or clusters (as shown in Fig.~\ref{fig:block}). We want a matching which allocates each node on the right side (a team) to nodes from different clusters on the left side (people). A set of edges is considered diverse if it connects left side nodes (people) from different clusters. For example, in Fig.~\ref{fig:block}, matching team $t1$ to person $p1$ and $p2$ is a non-diverse matching (as both $p1$ and $p2$ come from same color block), while matching it to  $p1$ and $p3$ is considered diverse. Note that the clustering can be pre-defined (like the country of origin of workers) or calculated using any attribute. The methods discussed here are agnostic to the choice of clustering method, and they assume that each item has a cluster label and we want to maximize coverage over different cluster labels. If the labels are country of origin of workers, then the optimal teams will have people from different countries.

We use a square-root-based diversity reward function which balances the number of nodes (e.g., people) selected from different clusters, adapted from the work of~\cite{Lin11:Class} on multi-document summarization. We first define some notations. $S_j \subseteq E$ is the subset of edges in a proposed matching that are also incident to team $j \in T$. Assuming people belong to $K$ clusters\textemdash\eg, of skillsets or levels of experience\textemdash $P_k \subseteq P$, $k \in [K]$ is a partition of all people $P$ (\ie, $\cup _k P_k ~=~ P$ and $P_k \cap P_{k'} ~=~ \emptyset$ for all $k\neq k'$). This means that each edge is associated with the cluster of the person it is incident on.
We also define $w_{i,j}$ as the quality (or expertise) of worker $i$ to do team $j$. In our context, for a specific team $j \in T$, we define an objective function $f_j : E \to \mathbb{R}$ which rewards diversity as follows:

\vspace{-0.3cm}
{\small
\begin{equation}
f(S_j) ~=~ \sum_{k=1}^K\sqrt{\sum_{\{i \ |\ i \in P_k \ \land \ (i,j) \in S_j\}} w_{i,j}}
\label{eq:eq_div}
\end{equation}
}
\vspace{-0.2cm}

The part within the square root function controls the quality such that a higher weight $w_{i,j}$ implies the person $i$ offers higher utility (better expertise or higher quality) for the job $j$. On the other hand, the sum of the square roots corresponding to each cluster means that adding nodes from the same cluster gives less marginal gain compared to adding nodes from a different cluster. Hence, it promotes diversity by preferring people from groups that have not been well represented in the teams so far.

Maximizing $\sum_{j \in T} f(S_j)$ over all legal matchings $S$ allows us to solve the offline diverse matching problem. To solve the offline problem, submodular function maximization techniques \cite{badanidiyuru2014fast} can be used; however, this assumes that we know exactly all of the people who will be available now and in the future. 
Note that we chose the objective function in Eq.~\ref{eq:eq_div} because it is submodular, it can be optimized using a mixed integer convex solver and has been shown to give a state-of-art performance in diversity measurement for document summarization tasks~\cite{lin2012learning}. However, there are other submodular functions too, which are used in literature to measure diversity (like Herfindahl Index~\cite{ahmed2019measuring}), and they can be used instead of Eq.~\ref{eq:eq_div}. The team formation algorithm, which we discuss later, can be integrated with any monotonic submodular function, for which we can estimate the optimal solution.

In the next section, we define the real-time variant of this problem where we do not assume to know exactly which people will arrive in the future and perform matching ``on-the-fly,'' which more accurately mirrors real-world team formation.

\section{Team Formation with Sequentially Arriving People}\label{sec:algos}
In our real-time model for team formation/assignment, we again model people and teams with a bipartite graph $G(P,T,E)$ where an edge $e = (i, j) \in E$ represents whether a person $i \in P$ can perform task or join a team $j \in T$. Teams are represented as the right side of the bipartite graph and people are considered on the left side. There is a firm with a limited budget of $B$ and a set of $N$ heterogeneous teams $T$ that need to form. People arrive one at a time from a large pool $P$. 
Each person $i \in P$ has a fixed cost $\cscreen{i}$ which is the cost of interviewing or screening the person, during which we learn their attributes (\eg, demographic information, skillset, \etc). 
After the interview/screener, the firm must either assign the person to one or more teams or reject the person. When a person is accepted for team $j$, she receives a payment/salary/bonus of $\cbonus{i}{j}$.
Note that while we mentioned using $b(i)$ to refer to the upper bound for any node $i$, to differentiate between the upper capacity of teams and people on the two sides of a graph, we use notations $R^+$ and $L^+$ also. Each team has an upper budget $R^+$ of the maximum number of workers it needs. Each person has an upper budget $L^+$ of the maximum number of teams she is willing to simultaneously participate in. Every time a person is interviewed/screened, the set of edges from the person to all teams is considered to ``arrive.''

Each person $i$ has a weight $w_{i,j}$ representing the local utility (\ie, fit, value, \etc) derived by the firm after matching her to $j$ (we assume that after team formation, the person performs the task). We use $M$ to denote the maximum number of people who can arrive, which is assumed to be known by the firm; typically, $M$ is determined by the firm's budget and screening cost $\cscreen{i}$.

With this setup, our problem can now be formulated as a real-time submodular maximization problem with $N$ knapsack constraints\textemdash the $N$ teams' upper bounds $R^+$.

\subsection{Overview of our streaming algorithm}
To perform real-time team formation, we treat people as a continuous \textit{stream}, and build upon past approaches to streaming algorithms to do diverse matching. Specifically, our objective function is monotonic submodular with an upper bound on the cardinality of people and teams. Recently proposed algorithms by \cite{7906050} attempts to solve the problem of real-time submodular maximization with $d$ knapsack constraints, for $d \in \mathbb{N}$ (fully described as Algorithm $4$ of \cite{yu2016streaming}). 
This algorithm estimates optima for the offline problem based on all items and then accepts or rejects edges based on feasibility and marginal gain being above a cut-off value. An optimum is estimated either using maximum possible marginal gain over all edges, or the current maximum marginal gain.

Algorithm by Yu \etal{}~\cite{yu2016streaming} cannot be practically applied to the team formation problem due to two reasons. 
First, it maintains multiple \emph{separate assignment solutions} and, when items arrive, they are accepted or rejected for each list separately.
An arriving item can be accepted by multiple lists and rejected by others. Practically, this would mean that when a person arrives at a firm, he or she is possibly allocated to several teams and rejected by others.
The person does their allocated job for all the teams they are accepted for and the firm maintains multiple possible allocations simultaneously.
After completing the real-time allocation phase (when all people have arrived), the firm would then ``select'' the allocation list that has with maximum utility. This would mean that many people previously allocated to (and already working on) teams would then be rejected. If a person has completed the task already, then their output gets wasted. Each person may have to be paid for all the tasks they did, while only a fraction of tasks is used.

Second, their algorithm has only capacity constraints, implying that in many situations, teams may receive fewer people than its upper bounds (due to strict filtering). This can be problematic in practical scenarios, where teams often require at least a minimum number of people and have upper bounds too\textemdash \ie, have both coverage and knapsack constraints.

In this paper, we address these two issues with modifications to algorithm of~\cite{yu2016streaming}, for practical team formation. We propose to use Algorithm~\ref{algo:change} for submodular maximization with $d$-knapsack constraints, where optima objective value ($OPT$) is known. 
In this algorithm, $c_{e,jj}$ is the cost of admitting an edge $e$ (corresponding to worker being allocated to team) for the $d$ constraints. For our case, with only maximum team size as capacity constraints, $c_{e,jj}$ is $1$ and the maximum capacity of a team equals  $b~=~R^+$. Running this algorithm requires an $\alpha$-approximation of the global optimum for the offline case, $\alpha \in (0,1]$.  $v$ is a value less than $OPT$ and greater than $\alpha OPT$, and we later explain how $v$ can be estimated. $\Delta~f(\emptyset, e)$ is the marginal gain of adding a single edge $e$ to a null set. $\Delta~f(S, {e})$ is the marginal gain of adding edge $e$ to the set $S$. This algorithm provides a $\frac{\alpha}{1+2d}$-approximation guarantee of the optimal solution, where $d$ is the number of knapsacks and $\alpha$ is the approximation factor up to which we can estimate the optima $\OPT{}$.

\begin{algorithm}
\DontPrintSemicolon 
\KwIn{v such that $\alpha \OPT{} \le v \le \OPT{}$, $\alpha \in (0, 1]$}
\KwOut{A feasible team allocation $S \subseteq E$}
$S \gets \emptyset$\;
\For{$i \gets 1$ \textbf{to} M}{
Find a permutation $\phi$ of all edges from $i$ s.t. $ \Delta f(S, \phi_1)  \ge \Delta f(S, \phi_2) \ge ... \ge \Delta f( S, \phi_N)$\\
\For{$e \gets \phi_1$ \textbf{to} $\phi_N$}{
	\If{ $c_{e,jj} \geq \frac{b}{2}$ and $\frac{\Delta f(\emptyset, e)}{c_{e,jj}} \geq \frac{2v}{b(1+2d)}$, for any $jj \in [d]$} {
    $S = \{e\}$; \textbf{return} $S$
}

\If{ $\sum_{l \in S \cup {e}} c_{l,jj} \leq b$ and $\frac{\Delta f({S, e})}{c_{e,jj}} \geq \frac{2v}{b(1+2d)}$, $\forall jj \in [d] $} {
    $S = S \cup \{e\}$ 
    }
 }
 }

\Return{$S$}\;
\caption{Real-time Diverse Matching}
\label{algo:change}
\end{algorithm}

We solve the problem of real-time team formation in three steps using Algorithm~\ref{algo:change}. First, we define a convex optimization problem and solve it to estimate an upper bound on $\OPT{}$. 
Second, instead of individual edges (items) arriving sequentially, we receive a batch of edges (corresponding to all teams a person could join) arriving together. We sort these edges for marginal utility provided by an edge and send them in decreasing order of marginal gain provided by them. By prioritizing tasks more suited to the skill set of a person, we improve the performance of our algorithm by giving strictly better results than random order. Third, we discuss setting $\alpha$ using marginal gains for clusters to guarantee that we can satisfy lower bounds too (given unlimited arrival of people). 
Note that in Algorithm~\ref{algo:change}, we have not explicitly mentioned the case with capacity constraints on people (when each worker cannot do more than $L^+$ jobs) or monetary budget constraints (when maximum budget $B$ is given for team formation), but adding these constraints is straightforward and does not change the algorithm. To add any additional constraints like budget or person capacity, we only need to define the individual cost incurred in selecting the corresponding node and the total budget allowed. For instance, considering the monetary case would mean cost $c_{l,jj}$ in Algorithm~\ref{algo:change} equals $\cbonus{i}{j}$ for the budget constraints and upper bound $b$ equals $B$. We do not model the screening cost $\cscreen{i}$ in accepting or rejecting a worker.

We provide a summary of the algorithm's intuition before diving into details on how to estimate parameters in it.
Algorithm~\ref{algo:change} decides the allocation for each edge (from a person to a task) independently. This means when a person arrives, it can do both~---~allocate the person to a new team or allocate the person to a team with existing qualified workers. Let us consider a simple case of three teams (T1, T2 and T3), maximum three team members in each team, and 15 people from three countries (A, B and C). Each person can be a part of maximum two teams. For simplicity of demonstration, we assume that everyone from all countries are equally good (unit weight). The optimal offline diverse solution should have three people from different countries allocated to each team.

Now, let us assume that people arrive in this order: {A1, A2, C1, B1, B2, B3, C2, C3, C4, A3, A4, A5, B4, B5, C5}.
When a person arrives, we first calculate how much marginal gain they provide to each team and decide the allocation of the team in descending order of marginal gain (Step 3 of Algorithm~\ref{algo:change}). The allocation will work as follows\footnote{The marginal gain of each person for their allocated task is shown in round brackets.}: A1 gets T1 ($1$) and T2 ($1$); A2 gets T3 ($1$); C1 gets T1 ($1$) and T2 ($1$); B1 gets T3 ($1$) and T1 ($1$); B2 gets T2 ($1$); B3 gets rejected by all tasks ($\sqrt{2}-1$) and finally C2 gets T3 ($1$). As all teams have received the required number of people, the rest are not needed. In this example, we assumed that we knew the offline optimal solution. In next section, we will explain how we can either estimate the optimal solution or circumvent the need of estimating the optimal solution by using the marginal gain values.

\subsection{Estimating the Optimum: Finding the maximum number of people from each cluster}

To estimate the optimum for the offline problem, we assume an unlimited stream of people exists, without knowing the number of people arriving from each cluster or their order. We make two assumptions. First, we assume that all people from the same cluster provide similar utility for any given team and, second, we assume that people are willing to participate in all teams. With these assumptions, we can formulate the diversity maximization problem for all teams by summing up submodular gains across each team and each cluster from Eq.~\ref{eq:eq_div}.
Let $y_{k,j}$ be the number of people from cluster $k$ matched to team $j$. Let $w_{k,j}$ be utility of a worker from cluster $k$ matched to team $j$.
The maximum number of people who can work in a given team is $R^+$. Hence we define the following problem:
 
 \vspace{-0.3cm}
{\small
\begin{equation}
\begin{aligned}
\underset{y}{\text{max}}
& \sum_{j=1}^N \sum_{k=1}^K\sqrt{w_{k,j} y_{k,j}}
& \text{s.t.}
& & \sum_{k=1}^{K}y_{k,j} \leq R_{j}^+ \;  \forall j \in [N]
\end{aligned}
\label{eq:eq_opt}
\end{equation}
}
\vspace{-0.2cm}

This is a concave maximization problem with linear constraints, and can be solved using a convex solver for real-valued $y$ and optimum value $\OPT{}^{*}$. A mixed-integer convex solver can also be used to obtain the true $\OPT{}$~\cite{lubin2016polyhedral}; however, such solvers are still in their nascency and, as we discuss later, the real-valued relaxation is sufficient for our case.

Solving Eq.~\ref{eq:eq_opt} with real valued $y$ yields $\OPT{}^{*}$, which satisfies $\alpha \OPT{} \le v \le \OPT{} \le \OPT{}^{*}$.
Solving this problem essentially estimates how many people from each cluster we should expect in an optimal solution and \emph{not} the allocation of individual people (as people are exchangeable within a cluster). We use $\OPT{}^*$ in place of $\OPT{}$ to filter edges in Algorithm~\ref{algo:change}.

Algorithm ~\ref{algo:change} accepts or rejects edges based on marginal gain and constraint satisfaction in Step 5. However, in practice, matching people to teams often also requires a lower bound of at least $R^-$ people for each team. In Algorithm~\ref{algo:change}, it is possible that the cut-off is too high for marginal gain (Step 5) and enough people do not get assigned to each team. To solve this problem, we pre-calculate the marginal gains for each cluster and find the $R^{-}th$ highest marginal gain among all clusters (denoted as $df_{R^-}$). This value is used to set the value of $v$ (used in Algorithm~\ref{algo:change}) such that:

\vspace{-0.2cm}
{\small
\begin{equation}
\begin{aligned}
& v \le  \frac{df_{R^-}.b.(1+2d)}{2}
\end{aligned}
\label{eq:setalpha}
\end{equation}
}
\vspace{-0.2cm}

Setting $v$ using Eq.~\ref{eq:setalpha} ensures that at least $R^-$ workers will get accepted by the algorithm irrespective of the arrival order of people as the marginal gain of ($R^-$)$^{th}$ person will still be below the cut-off in Step 5 of the algorithm. In the simulation results, we explain how setting $\alpha$ or $v$ not only helps ensure the lower bounds but also improves overall matching utility. 
If the optimization problem in Eq.~\ref{eq:eq_opt} is solved exactly with integral $y$, the current algorithm also provides a $\frac{\alpha}{1+2d}$ approximation of the optimal solution. The specific choice of $v$ or order of arrival of nodes does not alter the theoretical guarantees.
For clarity, we have provided a table with nomenclatures in the supplementary material [Insert link to Supplemental Material here.].

\subsection{Performance metrics for diverse allocation}
We measure the performance of diverse matching on two factors---how much cluster diversity it adds to the team and how much utility it loses for the requester relative to maximum-weighted matching. To measure improvement in diversity, we measure the Shannon entropy of a match for each team, with and without our method. Shannon entropy has been used to incorporate diversity in recommendations and matching \cite{DiNoia:2014:AUP:2645710.2645774,ijcai2017-6} and is also
widely used in the ecological literature as a diversity index. It quantifies the uncertainty in predicting the cluster label of an individual that is taken at random from the dataset. Entropy of a team is given by: $-\sum_{i=1}^K{(p_k \log{p_k})}$, where $p_k$ is the proportion of people on that team from cluster $k$. 
Hence, the impact of real-time diverse matching can be measured as improvement in average entropy for all teams. We define the \emph{entropy gain} (\EntropyGain{}) as:

\vspace{-0.3cm}
{\small
\begin{equation}
\EntropyGain{} = \frac{\mbox{Average entropy using a diverse matching rule}}{\mbox{Average entropy using baseline allocation}}
\end{equation}
}
\vspace{-0.2cm}

Entropy for a team is maximized if it has members with even coverage of different clusters; entropy is minimized when all people are from the same cluster. 

To measure the loss of utility due to diverse matching, we adopt the \emph{price of diversity} metric proposed by \cite{ijcai2017-6} which measures the trade-off in economic efficiency under a diverse matching objective. Specifically, we define two complementary versions of this metric. First, to measure the economic loss due to rejection of people by diverse matching, we define the price of diversity ($\PoDnum{}$) as:

\vspace{-0.3cm}
\begin{equation}
\small
\PoDnum{}=\frac{\mbox{Number of people interviewed for diverse allocation}}{\mbox{Number of people interviewed for baseline allocation}}.
\end{equation}
\vspace{-0.2cm}

For example, let us say a team requires four people and diverse matching rejects two people and finds an allocation after the arrival of the sixth person. If a baseline method accepts the first four people, $\PoDnum{}$ will be $1.5$, implying that encouraging diversity requires interviewing/screening $1.5$ times as many people. Normally, the cost of interviewing or screening candidates is low compared to the cost of the main team (\eg, paying their salary); thus, even large values of $\PoDnum{}$ may be acceptable, and will also depend on resultant entropy gain.

We also define utility-based price of diversity, $\PoDutil{}$, to measure the aggregate weight lost due to rejecting people by diverse matching as:

\vspace{-0.3cm}
\begin{equation}
\small
\PoDutil{}~=~\frac{\mbox{Utility obtained using baseline allocation}}{\mbox{Utility obtained using diverse allocation}}.
\end{equation}
\vspace{-0.2cm}

For example, say a team $j$ requires three people, and that people belong to one of three clusters $k \in \{1,2,3\}$ with team utilities $w_{\{1,2,3\},j} ~=~ \{3, 1, 1\}$, respectively. If we use a greedy algorithm as a baseline, it will maximize utility only by selecting people from the first cluster, accruing total utility of $9$, while diverse matching will accrue total utility of $5$ by selecting one people from each group. Hence, $\PoDutil{}$ will be $1.8$ against the greedy baseline.

\section{Experimental Results}\label{sec:experiments}
We begin this section by testing our algorithm on simulated results, showing how the price of diversity is affected by factors like how many people come from each cluster. Next, we deploy it on an online platform to show how filtering works in practice. We use our online platform to collect data from 50 online crowd workers, who are tasked to complete two tasks. Using this data, we then show our algorithm's performance on the true arrival order as well as the new unseen arrival ordering of workers.
While we compare our algorithm's final utility with the offline performance, we cannot use it as a baseline to calculate $\PoDnum{}$, as it requires all people to be present in a pool. Instead, we use the first-come-first-serve (random) allocation baseline for our experiments. For the baseline, people who satisfy all the constraints are allocated to tasks without optimizing for diversity. Lastly, we create another simulated dataset and show how the matching algorithm can be used to handle complex constraints and diversity for multiple attributes.

\subsection{Team formation for simulated agents}

In this section, we test our algorithm on simulated data. We consider simulated people sampled from different groups arriving in real-time and the algorithm assigns them to different teams. We demonstrate the effectiveness of our method in different situations when there are balanced or imbalanced clusters (or group identities), when the utilities of workers are different and when the arrival ordering of the workers varies.

For our study, we consider $10$ team tasks ($N~=~10$), each of which requires at most $3$ people ($R^+~=~b~=~3$). People are sampled from $3$ clusters ($K~=~3$).
In the real world, these clusters can be any label attached to a person, like the country of origin, race or area of expertise.
While the total number of groups is known beforehand, a person's group or cluster id is known only after she arrives (\ie, are interviewed/screened).
The cluster ID refers to any possible grouping of people. Clusters can be based on single attributes (like gender or country) or a combination of attributes.
Our model assumes that the utility obtained from all people sampled from the same group is the same. We start by simulating a situation where every person's utility is the same irrespective of what cluster they belong to, and all the clusters are roughly the same size. Next, we show what happens when people from particular clusters have higher or lower utility. We demonstrate how the parameter $\alpha$ affects matching performance in such cases. Finally, we show that our algorithm is robust, even for skewed distributions.

\subsubsection*{Clusters with equal utility}
Imagine a case, where a firm tries to recruit people who belong to three different professions. Each profession is valued equally to complete the task and roughly one-third of the applicants belong to each profession.
To model such a scenario, we consider three equally probable clusters offering equal utility, where all people have unit utility for all tasks, hence $w_{\{1,2,3\},j} ~=~ \{1, 1, 1\}$. We do not model the monetary cost of interviewing or total budget, so $c_{i,j}~=~1$ for all workers and teams. 
We do $100$ runs with a maximum of $100$ people $(M~=~100)$ streaming in random order. People are drawn from a multinomial distribution with cluster probabilities $\theta~=~[\frac{1}{3}, \frac{1}{3}, \frac{1}{3}]$, respectively.

Solving the optimization problem in Eq.~\ref{eq:eq_opt},  we find the offline optimal objective value $OPT^{*}~=~30.0$. For our simulation, we set $\alpha~=~1$ (which gives $v~=~30.0$). This gives the worst-case performance bound of $1.428$ for the real-time algorithm. Using Algorithm~\ref{algo:change} to filter edges, we obtained the team assignment for all the runs. In each run, we were able to find the optimal matching with utility $30.0$, which is also the offline optimal allocation (one person accepted from each cluster). Entropy for all teams in all $100$ runs is $1.09$, implying that all teams were formed with people from three different clusters. In our experiment, we find that, on average over the different runs, the median number of people we interviewed before forming a diverse team is $5$ people. For the run with the worst-case performance of our algorithm, we interviewed $8$ people and the run requiring minimum interviews had only $3$ interviews to hire $3$ people. 
Hence, median $\PoDnum{}$ is $2.67$, while $\PoDutil{}$ is $1$. This means that diverse matching improves coverage over clusters in all cases, but requires us to interview or screen $2.67$ times as many people before we can form diverse and high-quality teams. 
To facilitate reproducibility, we have provided a table with nomenclatures and the values used in the above experiment in the supplementary material.

\subsubsection*{Avoiding Task Starvation} 
In our optimization problem, we do not explicitly impose lower bounds (cover constraints) on teams or tasks, \ie{} we do not model a constraint saying each team must get a minimum threshold number of people. However, for real-time team formation, teams may require at least $R^-$ people to be effective. As discussed earlier, Eq.~\ref{eq:setalpha} can be used to set $\alpha$, guaranteeing the goal of meeting the minimum quota. Using Eq.~\ref{eq:setalpha}, $v \le 31.5$ for the previous case requiring $3$ workers for each task ($R^-~=~3$), so setting $\alpha~=~1$ satisfied this condition. 

The parameter $\alpha$ acts as a filter, as decreasing it lets the real-time algorithm accept more people from each cluster (forming less diverse teams for the sake of expediency) while increasing it accepts only the workers with highest marginal gain (holding out on candidates until it can form a diverse team). On one hand, setting $\alpha$ too high will mean most people get rejected, leading to a matching where a team never receives enough people. However, reducing $\alpha$ to very low values will essentially accept all people and behave similarly to random team formation (\ie, just allocate whichever person arrives first). For example, in the previous problem, when we reduce $\alpha$ to $0.4$, the median fitness drops to $24.14$, while the median entropy drops to $0.636$. This means that the median team has three workers, who belong to only two clusters. Hence, by smartly choosing the parameters of the algorithm, we can control how strict we want to be in the filtering of incoming workers.

\subsubsection*{Clusters with different utility}
Imagine a case, where a firm tries to recruit people who belong to three different professions. Each profession is valued differently with people from one profession being more desirable as another. Unequal weights can be allocated to people when those from a particular profession specialize in the task. Roughly one-third of the applicants belong to each profession.
To model such a scenario, we consider clusters with unequal cluster utility.  
In this work, we assume that we know the team task utility for each group after the screening task and that other methods like expertise identification can be used to identify how much a person is valuable to the team task at hand. In this simulation, we consider three clusters with utilities $w_{\{1,2,3\},j} ~=~ \{3, 2, 1\}$. 

\begin{figure}[t]
\centering
\begin{subfigure}[t]{0.4\textwidth}
\includegraphics[height=4cm]{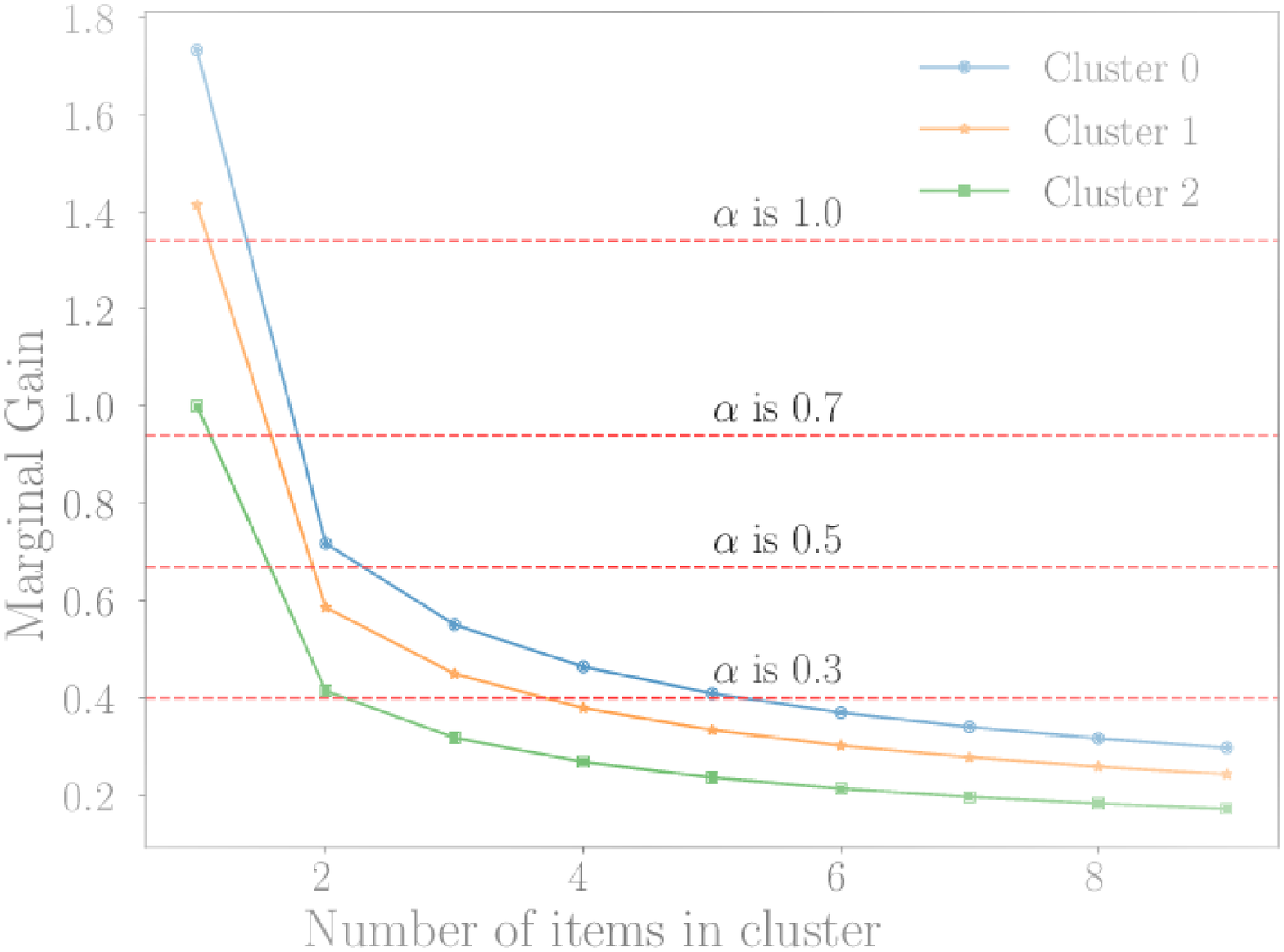}
\caption{Effect of $\alpha$ on worker acceptance from each cluster. \label{fig:alphab}}
    \end{subfigure}
    \begin{subfigure}[t]{0.4\textwidth}
\includegraphics[height=4cm]{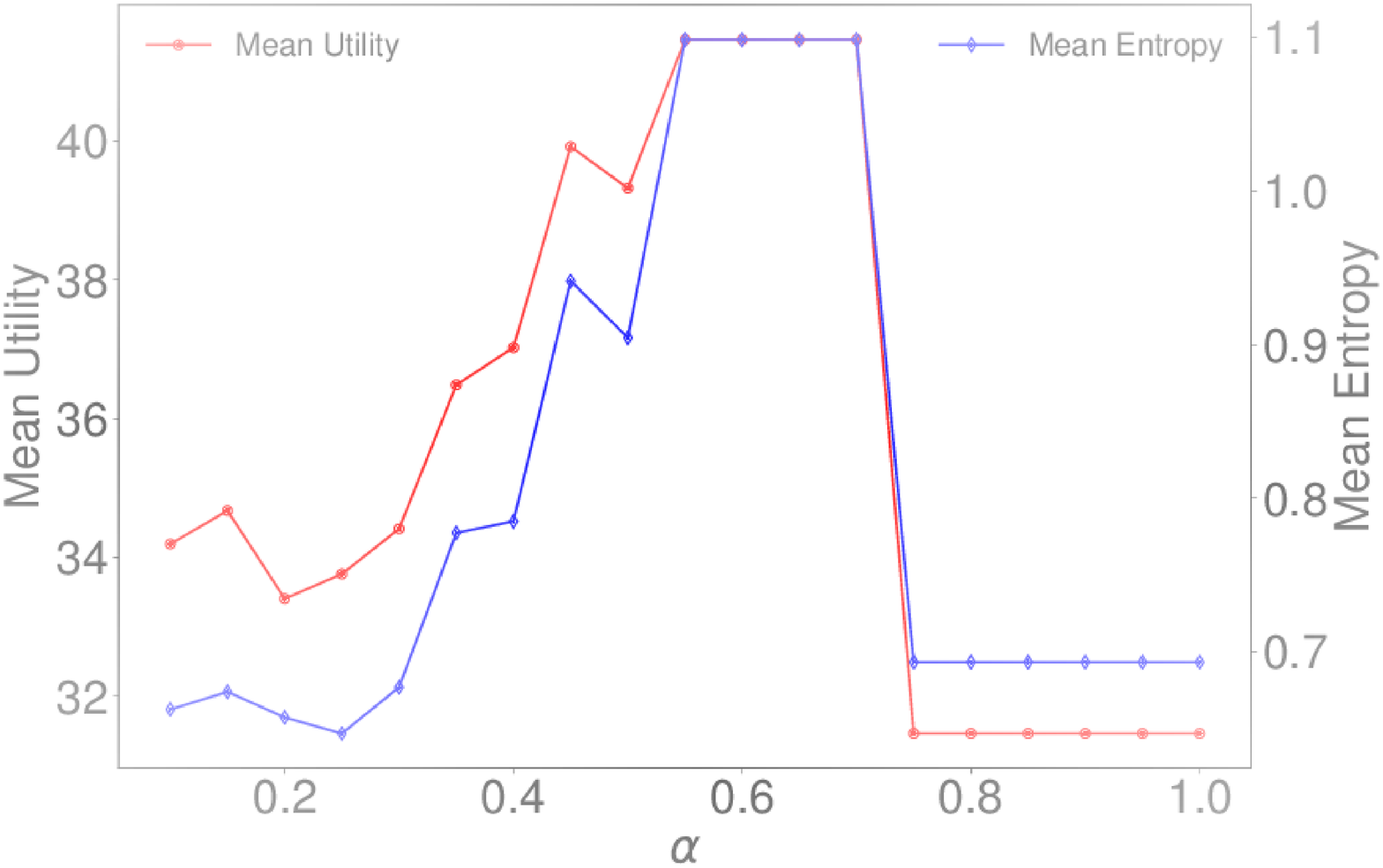}
\caption{Effect of $\alpha$ on utility and entropy. Utility maximizes when $\alpha$ is set according to 
Eq.~\ref{eq:setalpha} \label{fig:alpha}}
    \end{subfigure}
\vspace{-0.1in}
\caption{Effect of $\alpha$ on worker acceptance and mean utility
}
\end{figure}

On running the simulation, we found that setting $\alpha~=~1$ and simulating $100$ runs led to a median fitness of $31.4$ with all team tasks only matched to two people (one from cluster 0 and other from cluster 1). From Fig.~\ref{fig:alpha}, we notice that the marginal gain of the first person from Cluster 2 is $1.0$ (y value on the green curve corresponding to x=1). The red horizontal line for $\alpha~=~1$
has a y-intercept greater than $1.0$, hence this person will not be accepted by the algorithm.
The optimal fitness $OPT^*$ from Eq.~\ref{eq:eq_opt} is $42.42$. However, if $\alpha$ is reduced to $0.7$ (which is less than the cut-off of $0.74$ calculated using Eq.~\ref{eq:setalpha}), the desirable lower bound is met (each team receives three people) and the median fitness for $100$ runs improves to $41.46$ (which is also the optimum fitness for the offline problem). Hence, the real-time matching algorithm gives the optimum offline allocation of diverse teams.

In this case, the median entropy is $1.09$ with zero violations\textemdash \ie, all teams get three people from three different clusters. On average, the team forms after $5$ workers arrive. In the worst case, the team formed after $16$ workers arrived, leading to a median $\PoDnum{}$ of $1.67$.
Fig.~\ref{fig:alpha} shows how utility increases when lowering $\alpha$ initially, and then decreases on further reducing it. This is due to the submodular marginal gain of individual clusters as shown in Fig.~\ref{fig:alphab}. 
The x-axis shows the number of people selected from a single cluster for a single task. Here, each new person from a cluster provides less marginal utility and different clusters have different curves for marginal gain. In Step 5 of Algorithm~\ref{algo:change}, we accept or reject people if their marginal gain exceeds a cut-off directly proportional to $\alpha$ (as shown by the dotted red lines). We will accept people from a given cluster until the marginal gain curve for that cluster dips below the dotted line. The marginal gain of people belonging to each cluster is shown, where the first person from Cluster $0$ has a marginal gain of $1.73$ and the second person from Cluster $0$ has a marginal gain of $0.72$. Hence if $\alpha~=~1$, only a maximum of one item from Cluster $0$ will be accepted.
Similarly for Cluster $1$, if $\alpha~=~0.5$, a maximum of two people can be accepted.

Similarly, for $\alpha~=~0.3$, up to $5$ people from cluster $0$, $3$ people from cluster $1$ and $2$ people from cluster $2$ can be accepted. Although, the actual acceptance rate depends on the order in which people arrive, setting $\alpha$ less than $0.74$ guarantees that real-time diverse matching has zero violations as soon as one person from each cluster shows up. The theoretical lower bound on total utility, in this case, is $ \le 1.46$ and in practice, we get much better results. For the tasks requiring three workers, we first find the third-highest marginal gain among all clusters. As shown in Fig.~\ref{fig:alphab}, there are three clusters with weights 3 (blue), 2 (orange) and 1 (green). The top four marginal gain values are $\sqrt{3}, \sqrt{2}, 1 and \sqrt{6} - \sqrt{3}$. For $R^- ~=~ 3$, the third-highest value is $df_{R^-}~=~1$. After obtaining $df_{R^-}$, we calculate $v$ using Eq.~\ref{eq:eq_opt}.

\begin{figure}[ht]
\centering
\includegraphics[height=4.0cm]{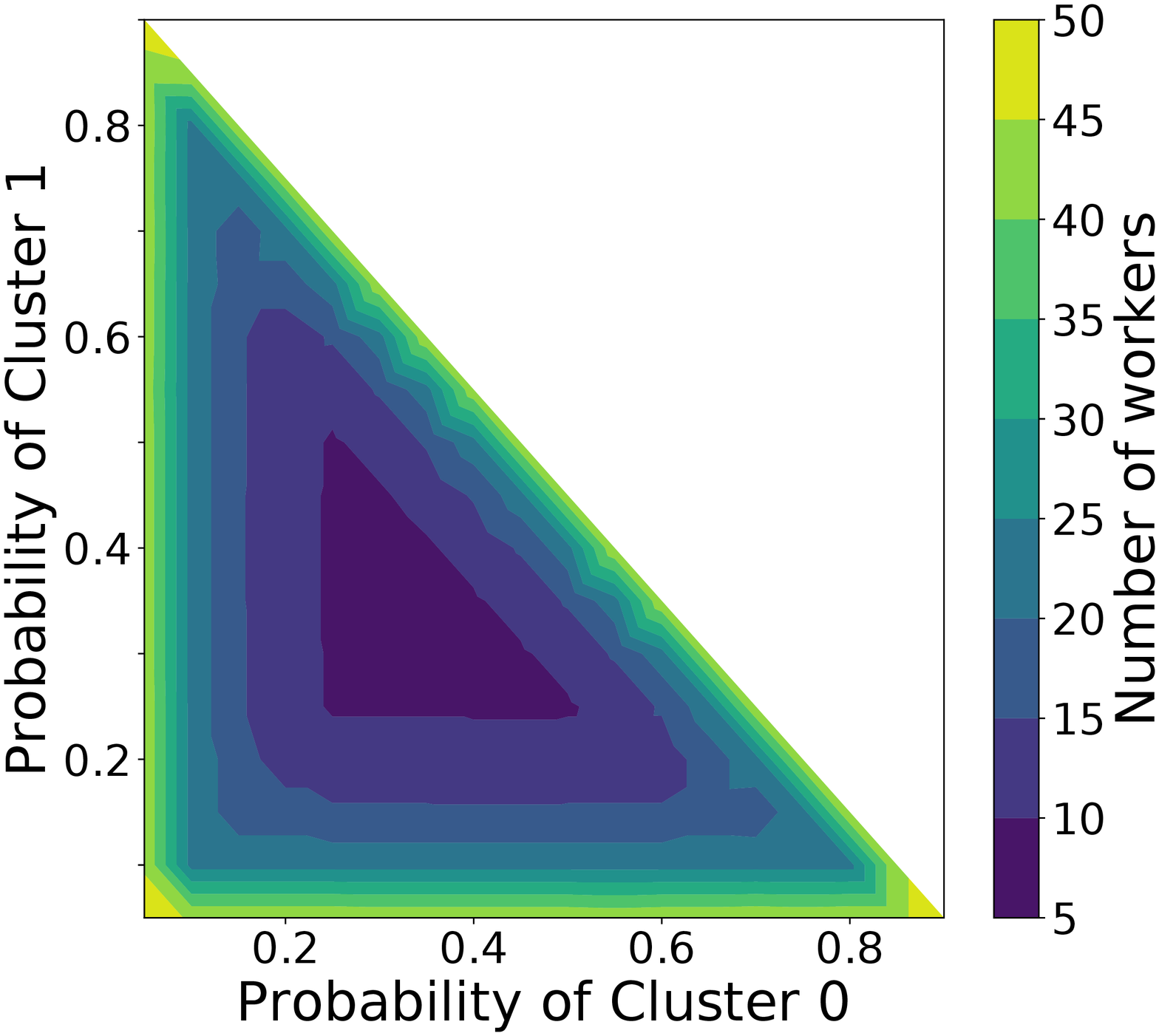}
\includegraphics[height=4.0cm]{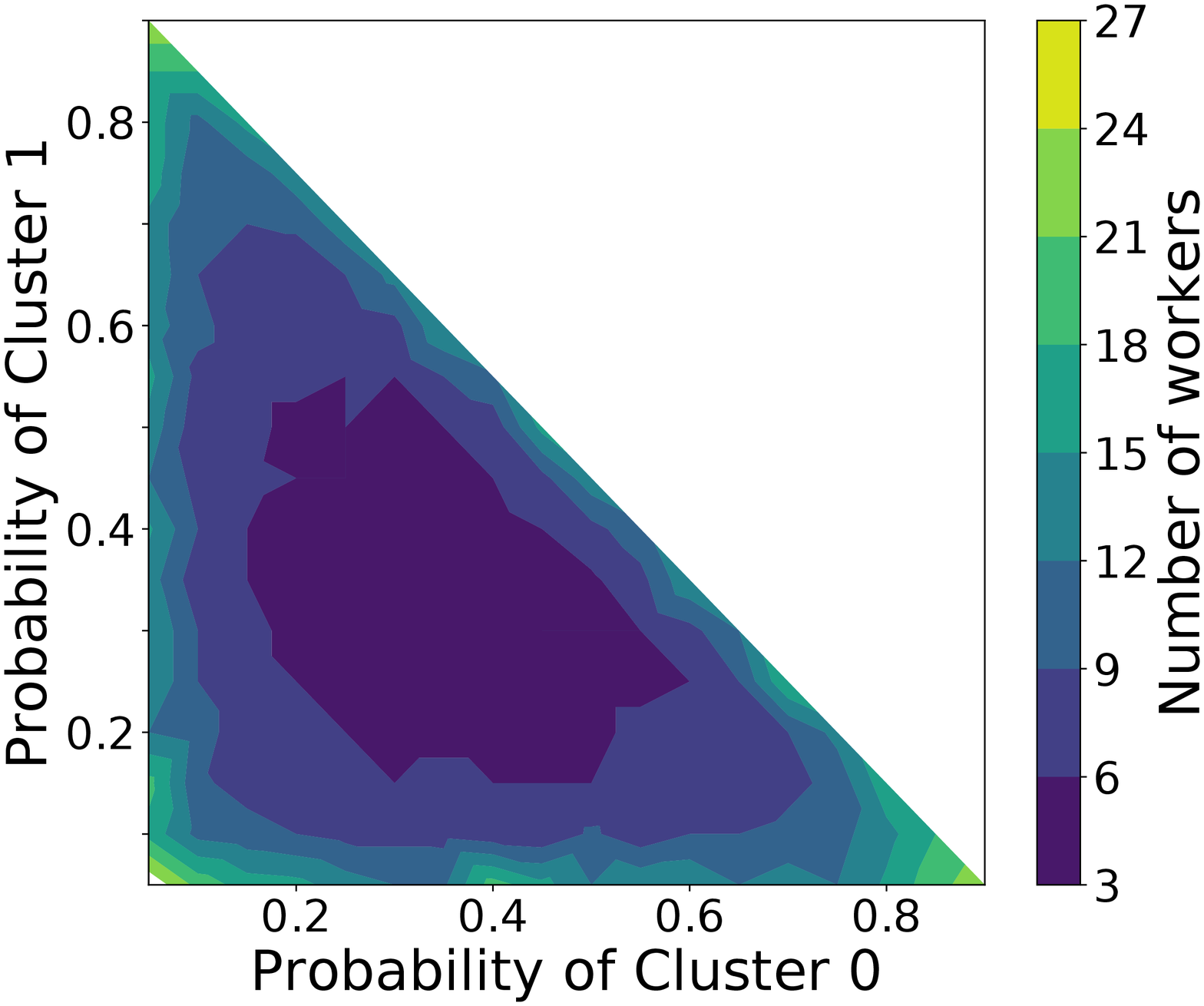}
\caption{Effect of Cluster Distribution. Left: Expected number of people needed . Right: Actual number of people needed (median of 100 runs).}
\label{fig:cls}
\end{figure}

\subsubsection*{Different sized clusters}

Imagine a case where a firm tries to recruit people who belong to three different countries. 
We assume that workers from different countries have different utility for different tasks and the number of workers from each country is different. Such a situation frequently occurs on Amazon Turk, when a person wants to assemble a team of people belonging to different countries from an online community or a pool of people (such as Mechanical Turk). If we consider three clusters being the US, India and all of the other countries, then past literature \cite{Difallah:2018:DDM:3159652.3159661, doi:10.1177/2158244016636433} has shown that approximately 75\% workers belong are from the US and 16\% from India. This means that if we draw randomly from the population, it is equivalent to sampling from a multinomial distribution with proportions $\theta~=~[0.75, 0.16, 0.09]$. Let us assume that the utility of assigning a person from these clusters to a team is  $w_{\{1,2,3\},j} ~=~ \{1, 2, 3\}$, respectively, implying a worker from India is twice more suitable for this task than a worker from the US.
If a firm knows these proportions, a natural and practical question to ask is, ``How much budget will I need to form a diverse team?''
or ``How many people should I expect to reject to form a diverse team?''

To answer this, we use the following example. A firm can only pay to interview at most $10$ people. When the firm starts interviewing, assume that $[6, 3, 1]$ people arrive from three clusters, respectively. As people are drawn from a multinomial distribution, we can calculate the probability of this event as:
$Pr(6, 3, 1)~=~$ $ \frac{10!}{6! 3! 1!}(0.75)^{6}(0.16)^{3}(0.1)^{0.09}~=~0.055$.
We also know the maximum number of people allowed from each cluster (\eg, one person), which means $7$ people will be rejected in expectation. 
Likewise, we enumerate all possible scenarios for different numbers of people coming from each group and calculate the expected number of people accepted for that distribution. In this case, we expect to accept $2.95$ people. This makes sense, as we need $3$ out of $10$ people to complete the task and in some cases, people may arrive only from one or two clusters. As we increase the number of people we interview, the expected number of accepted people also increases. Hence, we can calculate the expected number of people we need to screen to get $3$ people accepted for each team. 

Figure~\ref{fig:cls} shows the expected number of people needed to get the desired three people (zero violations) for different cluster probability distributions. The x-axis shows cluster 0's probability while the y-axis shows cluster 1's probability. Even for very skewed distributions with $\theta~=~[0.9,0.05,0.05]$, we get a $\PoDnum{}$ of only $15.4$. In plain words, this means that if 90\% of the worker population is from the US and only 5\% worker population is from India, and a firm wants to create a diverse team, it should expect to interview approximately 15 people for every one person accepted for the team. This is attributed to the skewed distribution, where people from certain clusters rarely show up.

In general, Fig.~\ref{fig:cls} is trying to demonstrate what happens when the number of people arriving from three different clusters is highly skewed--that this, how are my interviewing costs affected by diversity requirements, if there are (comparatively) few applicants in a given category? Each point on Fig.~\ref{fig:cls} is a distribution of people. The x-axis and y-axis show the proportion of people from the first two clusters. Let us say, we are trying to hire a team of three people, with people coming from three different countries (C1, C2, and C3). Now, we take a point on Fig.~\ref{fig:cls}, say {x~=~0.4, y~=~0.3}. This means 40\% of all the people are from C1 (\eg{} USA), 30\% of all the people are from C2 (\eg{} India) and remaining people are from C3 (\eg{} 30\% are from all the other countries). The dark blue color at {x~=~0.4, y~=~0.3} maps to a value of 5 people (as shown in the legend). As people from C1, C2 and C3 are in large proportions, on average, one only needs to interview five people to form a diverse team of three people. Hence, Fig. 4b shows how many people we have to reject before accepting a person for different proportions of populations.

In context, if people are paid \$1.00 to interview them compared to \$100.00 for doing the main task, then for zero expected violations (\ie, forming all teams) it costs only \$46.20 more compared to no screening and accepting the first three people---even under a highly skewed distribution of clusters with people from each of the two groups representing only 5\% of the population. In the median case, where distributions are more even, it only costs an extra \$5.00 to get a diverse allocation. Fig.~\ref{fig:cls} shows the results on simulating 100 runs for different probabilities of clusters and observing the median number of people needed by our algorithm.

\begin{table*}[!htbp]
\scriptsize
\centering
\caption{Distribution of various personal attributes in our MTurk experiment}
\resizebox{\textwidth}{!}{
\begin{tabular}{cc|cc|cc|cc|cc|cc}
\toprule
\shortstack{Age\\ ID} & Age & \shortstack{Gender\\ ID} & Gender & \shortstack{Education\\ ID} &  Education & \shortstack{Country\\ ID} &
Country & \shortstack{Politics\\ ID} & Politics & \shortstack{Race\\ ID} & Race\\
\midrule
0 & 18-24 (20\%) & 0 & Male (54\%) &  0 & \shortstack{High school degree \\ or equivalent (2\%)} &  0 & US (72\%) &  0 & Democrat (46\%) &  0 & White (56\%)\\
1 & 25-34 (48\%) & 1 &  Female (46\%) & 1 &  \shortstack{Some college credit,\\ No degree (12\%)} & 1 & India (28\%) & 1 &  Republican (30\%) & 1 &  Asian (30\%)\\
2 & 35-44(14\%) & & & 2 & Associates degree (12\%) & 2 & & 2 & Independent (20\%) & 2 & Hispanic (2\%)\\
3 & 45-54 (8\%) & & & 3 & Bachelors degree (50\%)  & & & 3 & Other (4\%) & 3 & \shortstack{American Indian or \\Alaska Native (6\%)} \\
4 & 55-64 (10\%) & & & 4 & Masters degree (22\%) & & & & & 4 & Other (6\%)\\
 & & & & 5 & Doctorate degree (2\%) & & & & & & \\
\bottomrule
\end{tabular}
}
\label{tab:demo}
\end{table*}

\begin{table*}[!htbp]
\centering
\caption{MTurk Price of Diversity ($\PoDnum{}$) and Entropy Gain in three cases: 1) Realized order (Column 2 and 3), 2) Median case (Column 4 and 5), and 3) Worst-case order (Column 6). We also report the additional expenditure of interviewing people for three positions in the realized order, when each interview costs \$0.10.}
\resizebox{0.8\textwidth}{!}{
\begin{tabular}{c|ccc|cc|c}
\toprule
Cluster & Entropy Gain & $\PoDnum{}$ & Additional Expenditure & Median Entropy Gain & Median $\PoDnum{}$ & Worst-case $\PoDnum{}$\\
\midrule
Age & 1.34 & 3.75 & \$1.12 & 1.23 & 1.25  & 7.25\\
Gender &  1.0 & 1.0 & \$0.30  & 1.33 & 2  & 10.5\\
Education & 1.33  & 2.0 & \$0.60 &  1.33 & 2.25  & 9.5\\
Country & 1.0  & 1.0 & \$0.30 & 1.23 & 1.5 & 9.5\\
Politics & 1.33 & 4.25 & \$1.27  & 2.0  &  3.75 & 12.25\\
Race & 2.0 & 10.75 & \$3.22  & 2.0 & 3  & 11.25\\
\bottomrule
\end{tabular}}
\label{tab:res}
\vspace{-0.2in}
\end{table*}

For clusters with different probabilities, we simulate $10$ teams and $100$ people, fix $\alpha~=~0.7$, and calculate the utility and entropy for $100$ different runs, drawing samples randomly according to the cluster $0$ and cluster $1$ probabilities. Each run randomizes the order in which people arrive. Our simulation shows that even for skewed distributions, our algorithm successfully finds high utility solutions. 
In all cases where people from all three clusters show up, real-time diverse matching finds solutions as good as the offline optimal solution. For edge cases, where not a single person from cluster $1$, $2$ or $3$ shows up, the competitive ratio (performance compared to the offline algorithm) is $0.81$, $0.79$ and $0.80$ respectively. In these edge cases, the team minimum requirement of workers (or lower bounds) is not satisfied by real-time matching as it only assigns two people per team rather than three. No one from the third cluster shows up and the algorithm does not accept multiple people from clusters which do show up to avoid a non-diverse allocation. However, out of the $171,00$ total orderings we simulated, only $40$ such violations occurred (\ie, teams did not get three people as nobody from one cluster ever arrived in only 0.23\% cases). 

We find that the median number of people needed for balanced distributions is low ($5$ people for $\theta~=~ [0.33,0.33,0.34]$). For skewed distributions, where people from one or more clusters rarely occur, the median number of people needed to be interviewed is more (27 people for $\theta~=~[0.05, 0.05, 0.9]$). The values are similar to the expected number of people shown in the left side of Fig.~\ref{fig:cls}, where we calculate the expectation values instead of simulating them.

\subsubsection*{Worst-Case Ordering}
So far, we have assumed that workers arrive randomly from a known or unknown distribution. However, imagine the worst case scenario, where workers come one at a time, in such an order that the cost of interviewing workers by the algorithm is maximized. Suppose a firm is willing to interview $20$ workers (some of whom they will hire), but it does not know how many people will come from each group. Assuming that the clusters have highly skewed utilities of $[1,30,30]$, that is workers from the second and third cluster provide 30 times utility compared to the utility of workers from the first cluster.
The optimal worker allocation is $[0,1,2]$ people from first, second and third cluster respectively, with $13.2$ utility  ($OPT^* ~=~ 13.52$).
We set $\alpha~=~0.75$ for zero violations, which means that the algorithm only accepts people from the last two clusters due to the skewed weights. However, the worst case ordering could have $20$ workers from the lowest weight cluster (cluster 0 in this case) all apply first. In such a case, the diverse matching strategy will not accept any of the first $20$ applicants. Hence, with a limited number of applicants, the algorithm can do arbitrarily bad if people from a few clusters never show up. In contrast, if an unlimited stream of workers is allowed, we are guaranteed to have no violations and will achieve a utility of $13.2$ when people from the second and third cluster eventually arrive. In the next section, we show that the price of diversity is not high in practice, even when workers arrive in the worst case ordering.

\subsection{Team formation for crowd workers arriving sequentially}
Using the simulation studies, we showed the efficacy of our real-time matching algorithm under different worker utilities and different probability distributions of classes. To further understand how the method performs when applied to a web platform to recruit workers, we conducted a crowdsourcing experiment. To test our algorithm for an online crowd team, we implemented diverse worker allocation on MTurk via two stages. We created a web platform, where we posted a screening task where people provided us demographic information. Next, we asked them to complete an ideation task in the second stage.
To make the experimental protocol easier and simpler to test and replicate, we selected writing tasks that were easy to complete in a short amount of time by team members and did not require the expertise of the workers in an Engineering domain. However, the methods developed in this work to filter workers are task-agnostic and can be applied equally well to relevant engineering tasks such as forming diverse technical teams or assigning design review tasks to diverse experts.

For the sake of demonstration, we assumed that our task required teams with education diversity, under the assumption that we wish to form teams with different educational backgrounds. Online crowd workers reported their educational background using six pre-specified categories ranging from ``High school degree or equivalent'' to ``Doctorate degree''. The categories and corresponding cluster id for various worker attributes are listed in Table~\ref{tab:demo}.
We categorized education up to a high school degree (ID 0) as Cluster 0, other non-graduate degrees (ID 1, 2, 3) as Cluster 1, and graduate degrees (ID 4, 5) as Cluster 2. In this task, we assumed that workers from Cluster 0 provide thrice utility compared to workers from Cluster 2.
This screening task filtered people using pre-set weights of $w~=~[3,2,1]$ for three clusters ($K~=~3$) and $\alpha~=~0.7$. 
It used two constraints $d~=~2$ corresponding to maximum people needed for each task.
We designed a platform, which after receiving a person's screener response, either directs them to the last page or allocates them to two different teams/tasks ($N~=~2$). 
Each team/task required 3 people ($R^+~=~R^-~=~b~=~3$), we paid $10$ cents for the screening task ($c_i^S~=~0.10$), and a $\$1.00$ bonus for the main task ($c_{i,j}^B~=~1.00$). When we started the experiment, we received people with education levels denoted by the following labels (ID's in Table~\ref{tab:demo}): $3, 1, 3, 1, 1, 4, 2, 3, 3, 3, 3, 4, 2, 3, 1, 3, 2, 0$. The first entry (3) shows that the first person indicated her educational level to be ``Bachelor's degree'' (from Table~\ref{tab:demo}), hence she belongs to Cluster 1, and so on for the remaining entries.

Upon running this experiment, we found that our algorithm accepted the first, second and eighteenth person, providing a diverse mix of education. Although the first three people could have provided a total utility of $6~(2+2+2)$, they all belonged to the same cluster and offered no diversity of educational level (zero entropy as first three people had a similar education level). Our algorithm's diverse allocation provided a utility of $7~(3+2+2)$. However, it incurred a cost of $\$4.80$ rather than the $\$3.00$ it would have paid for non-diverse allocation. $\PoDnum{}$ in this case is $6$ and $\PoDutil{}$ is $0.86$. The actual price of diversity in different situations depends on the order in which people arrive. 

To compare to counter-factual orderings, we ran another experiment where each person completed both tasks every time they accepted a job (\ie, we did not perform team formation immediately). This allowed us to measure each person's performance on all tasks. 
We then used this data to evaluate our algorithm by using the same data set to evaluate and compare several orderings/assignments.
We provided people with two questions to each participant, who had to submit their ideas on 1) ``How might we make low-income urban areas safer and more empowering for women and girls''
2) ``How might we restore vibrancy in cities and regions facing economic decline?''

These questions were selected as they are open-ended, complex and accepted different viewpoints. They did not require previous domain knowledge by the workers.
We ran the experiment in three batches ($M~=~50$ workers total). For the screening task, we requested demographics from each person regarding age, gender, education, country, political inclination~\footnote{While we used the terms ``Democrat'', ``Republican'', and ``Independent'' in our data collection, this is similar to ``Liberal,'' ``Conservative,'' and ``Moderate'' terms, respectively, used in other countries.}, and race. In general, we observed that the distribution of people in certain demographics was highly-skewed (\ie, most people belonged to one class)\textemdash see Table~\ref{tab:demo}.

Table~\ref{tab:res} lists the real-time matching results for three scenarios. 
First (column 2 and 3), we calculated the Entropy Gain and $\PoDnum{}$ for the actual order in which we received people. 
We considered six cases, corresponding to the six ways that individuals can be clustered (age, gender, education, country, politics and race). 
The results showed that we can achieve much higher entropy gain through diverse allocation compared to random allocation. For instance, the Entropy Gain to achieve diverse allocation for Politics is 1.33, while the $\PoDnum{}$ is 4.25. This meant that we gain in diversity but have to interview 17 people for every 4 people accepted in the team. Similarly, the $\PoDnum{}$ for Age, Gender, Education, Country and Race are 3.75, 1.0, 2.0, 1.0 and 10.75. 

While the realized order shows an instance of the performance of our algorithm, it is also possible that the next time we run it on an online platform, then people show up in a different order.
As the people we drew might not be representative of other possible orders, we took $1000$ permutations of those people and calculated how our algorithm performs in each case. Next, we calculate the median values for $\PoDnum{}$ and entropy gain (column 4 and 5). We notice that the real-time matching method successfully achieves large values for the median gain in entropy too. 
Finally, we calculate the worst-case scenario, where the people belonging to the smallest cluster show up last. As expected, $\PoDnum{}$ is higher but is not unreasonable due to the low cost of the screening task.


\subsection{Simultaneously Maximizing Diversity for Multiple Attributes}

\begin{figure}[!ht]
 \centering
 \includegraphics[height=8.5cm]{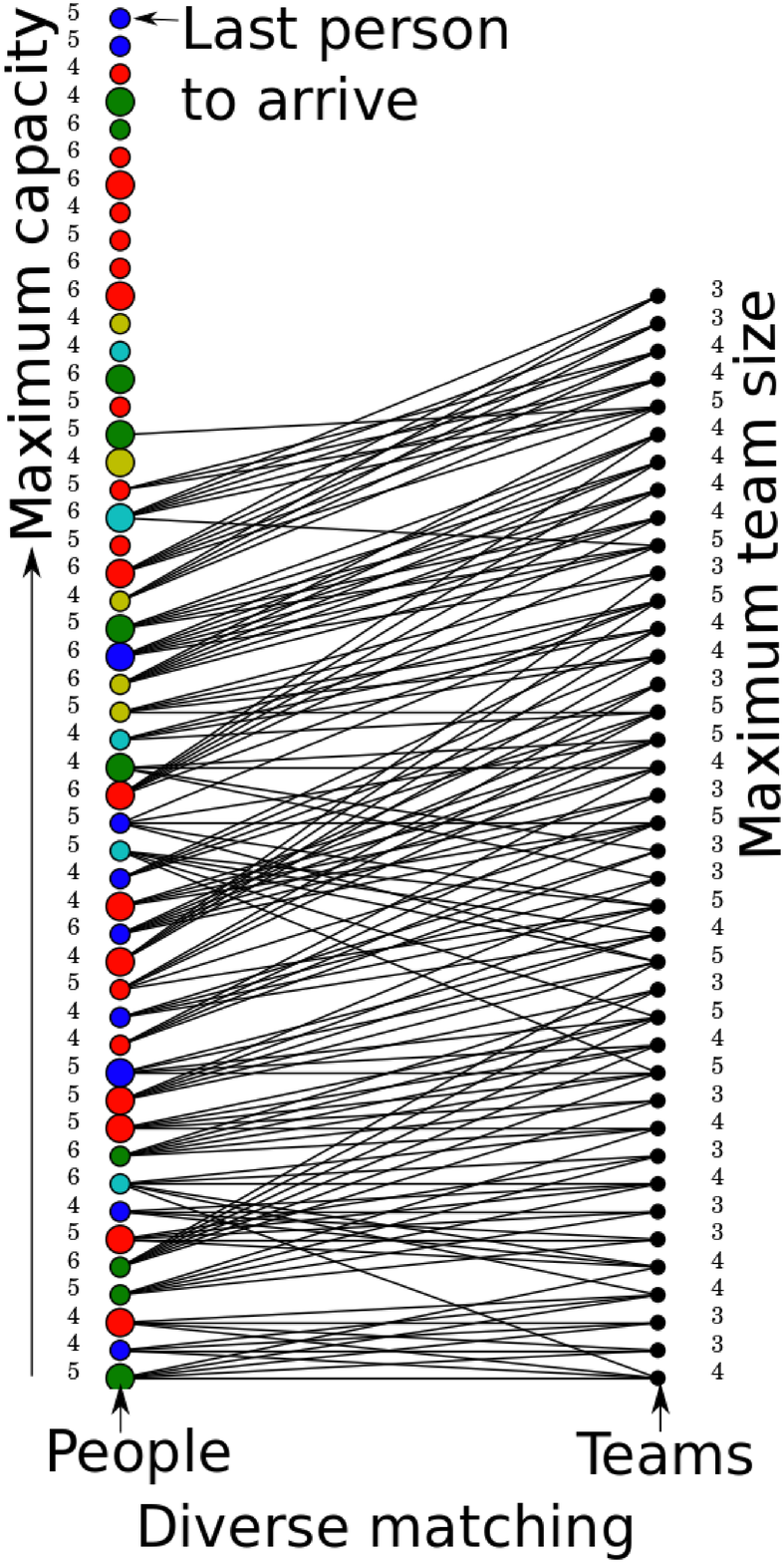}
  \includegraphics[height=8.5cm]{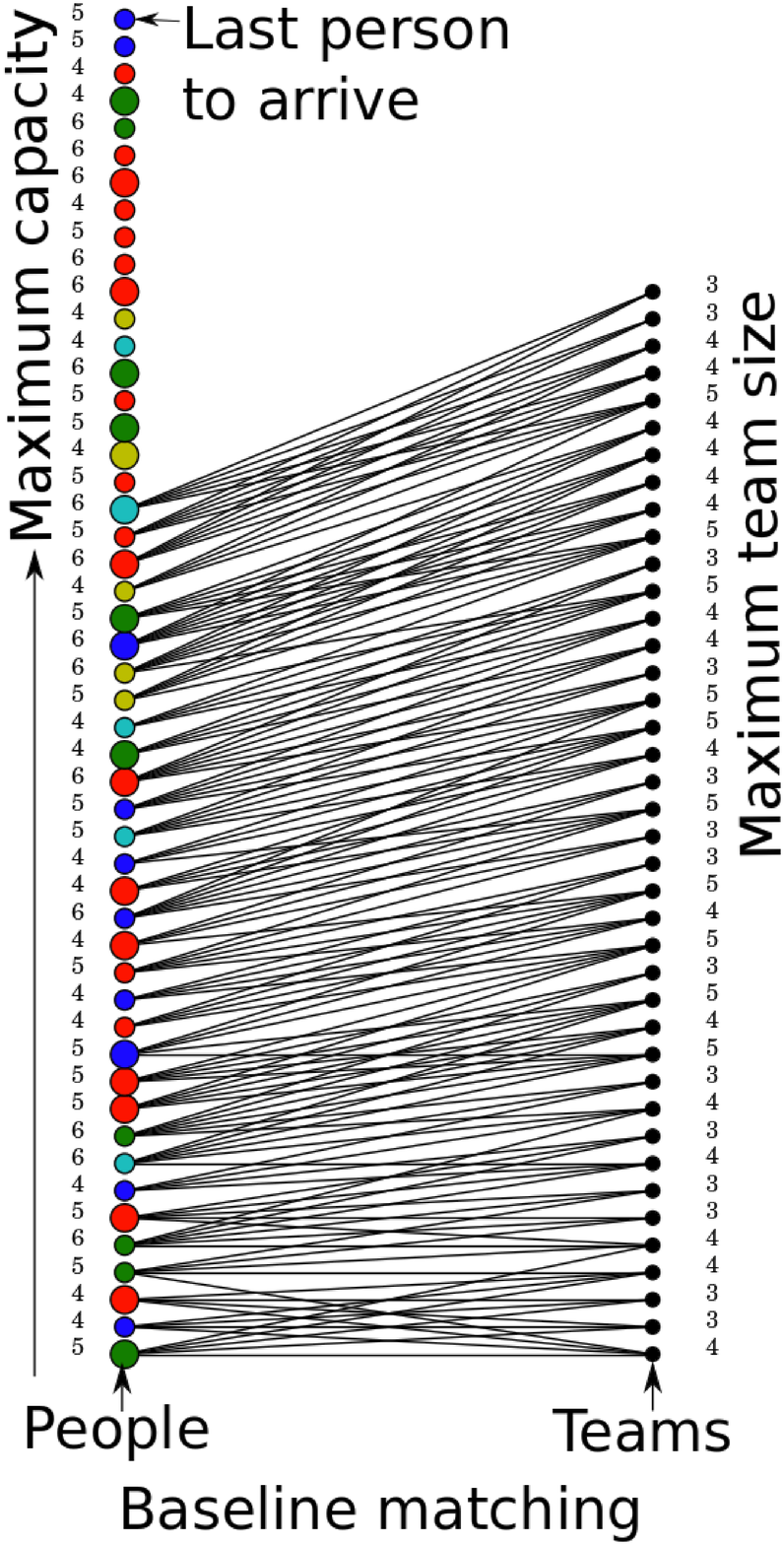}
 \caption{Bipartite illustration of the matching algorithm. People from different countries are shown by different colored circles on the left. Females are shown by larger circles compared to males. The number next to each person circle shows the maximum number of teams that they are willing to join and the number next to each team shows the maximum number of people needed in that team. The bipartite graph on the right side shows first-come-first-serve matching and the bipartite graph on the left side shows our diverse matching. We observe that for this particular order, diverse matching interviews only three extra people but leads to increase in gender entropy by 23\% and increase in country entropy by 18\%.}
 \label{fig:worker_capac}
 \end{figure}

In many real-world applications, one may want to allocate people to teams, such that teams are balanced for multiple factors like gender, skillset, experience, \etc{}
Our real-time diverse matching algorithm can also be used to form teams by simultaneously maximizing diversity based on multiple attributes. 

To demonstrate this, we experiment on a simulated example. Our goal of this experiment is two-fold. First, we want to show how we can modify the submodular objective function from Eq.~\ref{eq:eq_div} to measure diversity for multiple attributes (like gender and country of origin) simultaneously. Second, we also show that the algorithm can be used when different teams and different people have different demands. For example, one team may require three people, while another may have a demand of five people. People may also have different thresholds of the maximum number of teams they are willing to join. These additional requirements make the problem more difficult to compute, especially for a person trying to form teams manually. 

We start by defining a new objective function to maximize diversity for two attributes~---~gender and country of origin. Similar to Eq.~\ref{eq:eq_div}, we define a new objective function for a set $S_j$ of people matched to task $j$. We use $K_g$ to represent the number of unique genders and $K_c$ to represent the number of unique countries. $P_{kg}$ is the set of people who belong to gender $kg$ (For \eg{}, males are mapped to $kg=0$ and females are mapped to $kg=1$). We define, $P_{kc}$ as the set of people who belong to country $kc$. The objective function $f(S_j)$, measuring the quality and multi-attribute diversity of a team $j$ is defined as:

\vspace{-0.3cm}
{\scriptsize
\begin{equation}
f(S_j) ~=~ r~\sum_{kc=1}^{K_c}\sqrt{\sum_{\{i \ |\ i \in P_{kc} \ \land \ (i,j) \in S_j\}} w_{i,j}} +  (1-r)~\sum_{kg=1}^{K_g}\sqrt{\sum_{\{i \ |\ i \in P_{kg} \ \land \ (i,j) \in S_j\}} w_{i,j}}
\label{eq:eq_gen}
\end{equation}
}
\vspace{-0.2cm}

We use the weighing factor of $r$ to define the relative importance of gender and country diversity. Using Eq.~\ref{eq:eq_gen} above, we can calculate the objective value $f(S_j)$ of any team. By taking the difference between the objective values before and after adding a person to a team, we can calculate the marginal gain of that person for that team. This marginal gain is used in Step 7 of Algorithm~1 to decide whether a person gets allocated to that team or not. As $f(S_j)$ is a sum of two submodular functions, it is also submodular.

We create an experiment where there are 40 tasks and 50 workers. There are 14 tasks that require three workers, 16 tasks which require four workers and 10 tasks which require five workers, as shown by values next to teams in Fig.~\ref{fig:worker_capac}. There are 17 workers who will not accept more than 4 tasks, 18 workers who will not accept more than 5 tasks and 15 workers who will not accept more than 6 tasks. Similarly, the maximum number of tasks a worker is willing to accept is shown on the left of the worker nodes in Fig.~\ref{fig:worker_capac}. The figure has two bi-partite graphs, each representing the same people and teams, but different matching methods.
The left sides of the first bipartite graphs show nodes representing people and the right sides show teams. The number to the left of a person node shows how many tasks a person is willing to accept. The number to the right of a team node shows the maximum number of people that a team is willing to hire.

We assume that people belong to one of five possible countries. The countries C1 (red), C2 (blue), C3 (green), C4 (yellow) and C5 (cyan) have 20 workers, 10 workers, 10 workers, 5 workers, and 5 workers respectively. The left nodes in each bipartite graph are colored corresponding to their country of origin.  We also assume that each person belongs to one of two possible genders. $60\%$ of the workers are male and the remaining $40\%$ workers are female. The left nodes corresponding to the female gender are double the size of male nodes in the graphs. It is important to note that when simultaneously maximizing diversity for two different attributes, these diversities may conflict. A newly arrived person may increase the gender diversity of the team but not the country diversity, while another newly arrived person may add to the country diversity and not gender diversity. In our experiments, we weight these two factors equally by setting $r=0.5$. We also set the edge weights of all workers to all teams to one, implying that people from all countries and gender are equally good for all teams.

For people arriving sequentially, one cannot judge an algorithm based on a single permutation, as it is possible that for a particular sequence of people the algorithm can perform very well or poorly. For instance, if one male and one female arrive alternately, a first-come-first-serve algorithm will also give the most diverse matching for gender. To compensate for differences in arrival order, we conducted 100 runs with different permutations in which people arrive. We measure the performance of our algorithm using Gain in Entropy (GiE) and Price of Diversity (PoD) metrics defined before for each run. The Gain in Entropy (GiE) for gender ($GiE_g$) and country $GiE_c$ are defined separately, to measure improvements in diversity for both types of attributes. As a baseline, we use a first-come-first-serve algorithm, which allocates people to teams by only satisfying the constraints.

For 100 runs, our results show that the random allocation gives an average gender entropy of $0.521$, while average country entropy is $0.906$. Average gender entropy of $0.589$ and an average country entropy of $1.119$ is obtained from our algorithm, which are large improvements over the baseline. Note that these averages are of 100 runs, where each run has 40 teams in them (effectively, it is an average of forming 4000 teams). We observe that our algorithm gets more diverse teams for both gender (average $GiE_g~=~1.130$) and country (average $GiE_c~=~1.235$).  The average Price of Diversity (PoD) of 100 runs is $1.143$. This means that if the baseline algorithm interviewed 100 people to do the allocation, on average the diverse algorithm interviewed 14 more people. To explain the results, we next show the improvement for an individual run (selected randomly). 

We consider the permutation of arrival order shown in Fig.~\ref{fig:worker_capac}. One can note from the differences between baseline and diverse allocation that the number of people interviewed for baseline was $32$ and the number of people interviewed for diverse matching was $35$ (the $PoD~=~\frac{35}{32}~=~1.093$). Both cases satisfied all the constraints and met all team demands. For this run, the gender entropy and country entropy of the baseline algorithm are $0.473$ and $0.977$. Using our diverse matching algorithm, the gender entropy is $0.581$ and the country entropy is $1.150$.
The baseline allocation had 9 teams (out of 40) which had all people from different countries. In contrast, our allocation had 21 teams which had all members from different countries. The baseline allocation had 9 teams where all people were of the same gender (non-diverse), while our allocation had only 3 such teams. Hence, by interviewing just three extra people, our algorithm led to large improvements in both gender and country diversity. This shows the efficacy of diverse matching on multiple attributes and complex constraints. 

Using this complex setting, we showed that our algorithm can show large improvements over a baseline algorithm. While we did not conduct an additional experiment comparing the algorithm against a human, manually trying to form teams, we argue that forming a diverse team for multiple attributes (gender and country) and satisfying the constraints for all workers and teams is a difficult task for a person to do manually within a reasonable amount of time. Even if a person can find a good solution manually, the process will not be efficient or scalable. For instance, we conducted 100 runs for different arrival orders of people, effectively forming 4000 teams (15,600 total allocations) within a few minutes. We can also scale up our experiment to include a much larger example, doing millions of allocations and incorporate different utilities. These qualities make algorithmic team formation a necessity in simultaneously forming multiple diverse teams.

 \section{Discussion}\label{sec:discussion}
Our above algorithms provide a scalable way to perform real-time, diverse, team formation that mirrors some of the constraints of real-world collaborative work and teams. However, our work leads to many open questions: 1)~What kinds or types of diversity is our approach well- or ill-suited to include? 2)~When in collaborative team formation would one want real-time diverse formation versus not? And 3)~what kinds of diverse team formation tasks or constraints would limit the approach we outline here?

\subsection{Handling different types of diversity}
Our above results demonstrated how to form diverse teams which were diverse with respect to people who were clustered into discrete groups (in our case, specifically based on demographics). We also showed that the method is generic in the sense that it can be easily applied to any type of diversity wherein people can be categorized into a set of groups\textemdash whether it is based on demographics, task-related skills, cognitive preferences, \etc~ 
In the supplementary material [Insert link to Supplemental Material here.], we added an additional experiment to show how our algorithm can be applied to a real-world application of allocating reviewers to sequentially arriving journal papers. This demonstrated that the algorithm is also applicable when the sequentially arriving side is teams. However, there are two important cases that we do not explicitly handle above: 1) where people can belong to multiple groups/clusters (\ie, where the clusters are not mutually exclusive) and 2) where there are not discrete clusters but rather continuous scales or spectra along which people vary.

When people may belong to multiple, non-mutually-exclusive clusters, one must modify our objective function in Eq.~\ref{eq:eq_opt} to consider not just the given weight assigned to that individual's group-to-team edges, but also other edges from other groups that the person may belong to. For instance, a person may have political affiliation as $50\%$ Democrat and $50\%$ Republican. If such a person gets matched to a team which tries to maximize the diversity of political views, then all both groups get credit proportional to the percentage membership of the person.
This increases the computational cost slightly (in that we have to consider more edges), but does not substantively change the above algorithm or results.

When people are mapped to a continuous or ordinal spectra (\eg, right-to-left leaning, \etc) rather than in groups (\eg, Democrat or Republican, \etc), diversity is often cast as a type of area, volume, or density coverage over a space. This changes the objective function\textemdash for example, using Determinantal Point Processes~\cite{kulesza2012determinantal} instead of entropy over groups. In such cases, our greedy algorithm remains the same so long as the coverage function is submodular, but estimating OPT is more challenging. Methods for doing so are a fruitful area for future research.

\subsection{Under what conditions would one want diverse team selection?}
Theoretically, our proposed method applies to any situation where people belong to different groups and we want even coverage of those groups (\eg, in team membership). However, practically, there are two important factors to consider. 
First is the price of encouraging diversity, especially in skewed distributions. In our simulated and human experiments, when some of the clusters or groups were quite rare, it was possible that requiring diverse matching rejected many people (while waiting for a person from a rare group to arrive). This rejection can have a non-trivial cost (\eg, when interviewing people), which may affect the total budget. In such cases, one must balance the cost of rejection with the skewness of the applicant pool. If the cost of rejection is high or there are few applicants from a given cluster/group, then diverse matching can become expensive. In some situations, however, this higher cost may be worth the commensurate benefits of a diverse team. 

Second, understanding that benefit-cost trade-off is central to knowing when and how to apply automated diverse team formation.  Diversity is often portrayed as a ``double-edged sword'' in contemporary organizational theory \cite{williams1998demography}. At one end of the spectrum, proponents stress how heterogeneity helps team outcomes, while opponents posit that heterogeneous teams may lead to dysfunctional interactions or sub-optimal performance. Different researchers who study collaborative work have looked at diversity from the lens of creative output \cite{siangliulue2015toward, kim2014ensemble}, team satisfaction \cite{ye2017does} or tie formation \cite{dong2016embracing} \etc~Although teams are routinely assembled from individuals with varying degrees of demographic and cognitive abilities, it is still an open question as to under what conditions heterogeneous composition leads to groups which outperform homogeneous teams \cite{horwitz2007effects}. While the answers to those questions lie beyond the scope of this paper, our proposed method complements existing research on the benefits of diversity by allowing one to mathematically study whether balancing one type of diversity might be useful for a domain. For example, by calculating the ``price of diversity,'' our method helps researchers in quantifying the impact of diversity on real-time team formation or other real-time matching problems.

As an example, consider two tasks. Task 1 requires a team to craft policies for an important national issue, while Task 2 requires the team to jointly write a review for the movie ``Titanic.'' Assume that the manager wants to maximize diversity with respect to political affiliation (Democrats, Republicans, Independent, Others) for these two teams. As in our simulation studies, one can use population estimates to calculate the expected price of diversity. For instance, we observed a $\PoDnum$ of $4.25$ on Amazon Turk. This means, to form a team of $4$ people for this task, we expect to reject another $13$ people. 
Getting this estimate and comparing it to a firm's costs and internal values illuminates the pros or cons of political affiliation diversity in each team. For the first task, opinions from diverse political viewpoints will make the policy stronger and may be worth the rejection costs. On the other hand, current research does not indicate that political diversity substantially benefits dramatic movie review writing, and thus may not be worth the rejection cost. In such cases, the firm can decide whether more research is needed to establish the benefit or not. Our method can be adapted to estimate the trade-off between the total cost of team formation and the utility gained by forming diverse teams.

\subsection{Limitations of diverse team formation}
From the simulations provided, one may wonder why a computational method is needed at all. Can diverse matching just be done manually? For a small number of teams and clusters, where all team members are equally qualified for the tasks, it is possible to form diverse teams manually. However, when the constraints are more complex (\eg, different tasks have different demands, multiple clusters exist, and different people have different utility) it quickly becomes impossible for a human to select diverse teams. In such cases, our diverse team formation method applies.

Another important implication of our research lies in a better understanding of team member utility. In our simulations, we assumed that we already knew the edge weights or the utility that a person offers to all the tasks. In practice, it is non-trivial to estimate that utility and a large body of research have looked into estimating a person's task utility~\cite{allahbakhsh2013quality}. Future research directions can look at this problem holistically, to estimate utility for diverse teams. One interesting direction would be integrating real-time diverse team formation with simultaneous utility assessment (\eg, based on worker accuracy in crowd markets).

Likewise, one must estimate a person's cluster or group. This paper used demographic groups but our method allows groups based on any factor. With some modification to the objective function, it is possible to allow multiple group membership too. However, defining groups in itself are non-trivial for some applications, and a person's group, affiliations, or characteristics may change over time. These questions complement our line of work and would be interesting areas for future research.

\subsection{Extensions beyond team formation}
Thus far, we have discussed how to form diverse, collaborative teams. However, team formation can benefit from diversity in two different ways\textemdash by joint team effort or just by aggregating individual efforts. For the former, organizational research has investigated many factors where diversity may benefit team output. However, a less obvious application of diverse team formation is the scenario where the team members work independently. In such cases, one expects to benefit from aggregating their individual outputs to form a collective output. Conference or journal paper reviewing is one example of this situation, where reviewers are not necessarily collaborating together, but aggregating reviews from diverse viewpoints will benefit a paper more than those from the same viewpoint. Diverse matching also applies to such broader definitions of team tasks. For instance, many online design communities expect participants to also review and critique each others' designs~\cite{fuge:openideo_collaboration,fuge:idetc_2014_oi}. By matching diverse sets of individuals to each design, one can expect to get reviews from different viewpoints. Real-time matching is necessary in this case as people arrive randomly over time and need a subset of designs to review. Similar issues arise in network science and formation as well, such as the preferential attachment problem.

\section{Conclusions \& Future Research}\label{sec:conclusions}

We presented an algorithm for assigning sequentially arriving people from different groups to teams\textemdash real-time diverse matching. We show that by using a low-cost screening task, one can group people and then allocate them to teams as they arrive while balancing the team diversity.
While we clustered people into groups based on demographics, our method is generic and can be applied to other attributes like expertise. Our method also applies to other real-time allocation tasks where diversity of viewpoints might matter: \eg, real-time worker-to-team assignments, journal paper-reviewer assignments, and intelligence analysis tasks.
Future work could include: 1) journal paper-review assignments where both the static and dynamic side of the bipartite graph are clustered; 2) latent or non-mutually exclusive cluster labels/attributes; and 3) combining real-time diverse matching with real-time cluster identification using Bayesian techniques~\cite{moreno2015bayesian}.

 \section*{Acknowledgements}
 The authors thank the anonymous reviewers for their input which significantly strengthened the manuscript. Ahmed and Fuge acknowledge partial financial support through NSF CMMI-1728086. Dickerson acknowledges partial support by NSF CAREER Award IIS-1846237, DARPA SI3-CMD Award S4761, and a generous gift from Google.

%
% ...

\bibliographystyle{asmems4}


\bibliography{MD-19-1496-references}

\begin{thebibliography}{10}

\bibitem{stirling2007general}
Stirling, A., 2007.
\newblock ``A general framework for analysing diversity in science, technology
  and society''.
\newblock {\em Journal of the Royal Society Interface, \textbf{ 4}}(15),
  pp.~707--719.

\bibitem{hewlett2013diversity}
Hewlett, S.~A., Marshall, M., and Sherbin, L., 2013.
\newblock ``How diversity can drive innovation''.
\newblock {\em Harvard Business Review, \textbf{ 91}}(12), pp.~30--30.

\bibitem{hunt2015diversity}
Hunt, V., Layton, D., and Prince, S., 2015.
\newblock ``Diversity matters''.
\newblock {\em McKinsey \& Company, \textbf{ 1}}, pp.~15--29.

\bibitem{paulus2016cultural}
Paulus, P.~B., van~der Zee, K.~I., and Kenworthy, J., 2016.
\newblock ``Cultural diversity and team creativity''.
\newblock In {\em The Palgrave Handbook of Creativity and Culture Research}.
  Springer, pp.~57--76.

\bibitem{Horton17:Effects}
Horton, J.~J., 2017.
\newblock ``{The Effects of Algorithmic Labor Market Recommendations: Evidence
  from a Field Experiment}''.
\newblock {\em Journal of Labor Economics, \textbf{ 35}}(2), pp.~345--385.

\bibitem{Kurata15:Controlled}
Kurata, R., Hamada, N., Iwasaki, A., and Yokoo, M., 2017.
\newblock ``Controlled school choice with soft bounds and overlapping types''.
\newblock {\em Journal of Artificial Intelligence Research, \textbf{ 58}},
  pp.~153--184.

\bibitem{Drummond15:SAT}
Drummond, J., Perrault, A., and Bacchus, F., 2015.
\newblock ``Sat is an effective and complete method for solving stable matching
  problems with couples.''.
\newblock In Proceedings of the International Joint Conference on Artificial
  Intelligence (IJCAI), pp.~518--525.

\bibitem{Charlin13:Toronto}
Charlin, L., and Zemel, R.~S., 2013.
\newblock ``The {T}oronto paper matching system: an automated paper-reviewer
  assignment system''.
\newblock In International Conference on Machine Learning (ICML).

\bibitem{Liu14:Robust}
Liu, X., Suel, T., and Memon, N., 2014.
\newblock ``A robust model for paper reviewer assignment''.
\newblock In ACM Conf. on Recommender Systems (RecSys).

\bibitem{Bertsimas13:Fairness}
Bertsimas, D., Farias, V.~F., and Trichakis, N., 2013.
\newblock ``Fairness, efficiency, and flexibility in organ allocation for
  kidney transplantation''.
\newblock {\em Operations Research, \textbf{ 61}}(1), pp.~73--87.

\bibitem{Dickerson15:FutureMatch}
Dickerson, J.~P., and Sandholm, T., 2015.
\newblock ``{FutureMatch}: Combining human value judgments and machine learning
  to match in dynamic environments''.
\newblock In AAAI Conference on Artificial Intelligence (AAAI), pp.~622--628.

\bibitem{Benabbou18:Diversity}
Benabbou, N., Chakraborty, M., Xuan, V.~H., Sliwinski, J., and Zick, Y., 2018.
\newblock ``Diversity constraints in public housing allocation''.
\newblock In International Conference on Autonomous Agents and Multi-Agent
  Systems (AAMAS).

\bibitem{cummings2009organization}
Cummings, T., 2009.
\newblock {\em Organization development and change}.
\newblock Wiley Online Library.

\bibitem{clutterbuck2011coaching}
Clutterbuck, D., 2011.
\newblock {\em Coaching the team at work}.
\newblock Nicholas Brealey Publishing.

\bibitem{lenhardt2004coaching}
Lenhardt, V., 2004.
\newblock {\em Coaching for meaning: The culture and practice of coaching and
  team building}.
\newblock Insep Editions.

\bibitem{levi2015group}
Levi, D., 2015.
\newblock {\em Group dynamics for teams}.
\newblock Sage Publications.

\bibitem{holpp1999managing}
Holpp, L., 1999.
\newblock {\em Managing teams}.
\newblock McGraw Hill Professional.

\bibitem{humphrey2009developing}
Humphrey, S.~E., Morgeson, F.~P., and Mannor, M.~J., 2009.
\newblock ``Developing a theory of the strategic core of teams: a role
  composition model of team performance.''.
\newblock {\em Journal of Applied Psychology, \textbf{ 94}}(1), p.~48.

\bibitem{sassenberg2007some}
Sassenberg, K., Jonas, K.~J., Shah, J.~Y., and Brazy, P.~C., 2007.
\newblock ``Why some groups just feel better: The regulatory fit of group
  power.''.
\newblock {\em Journal of Personality and Social Psychology, \textbf{ 92}}(2),
  p.~249.

\bibitem{barrick1998relating}
Barrick, M.~R., Stewart, G.~L., Neubert, M.~J., and Mount, M.~K., 1998.
\newblock ``Relating member ability and personality to work-team processes and
  team effectiveness.''.
\newblock {\em Journal of Applied Psychology, \textbf{ 83}}(3), p.~377.

\bibitem{jordan2002workgroup}
Jordan, P.~J., Ashkanasy, N.~M., H{\"a}rtel, C.~E., and Hooper, G.~S., 2002.
\newblock ``Workgroup emotional intelligence: Scale development and
  relationship to team process effectiveness and goal focus''.
\newblock {\em Human Resource Management Review, \textbf{ 12}}(2),
  pp.~195--214.

\bibitem{ross2010crowdworkers}
Ross, J., Irani, L., Silberman, M., Zaldivar, A., and Tomlinson, B., 2010.
\newblock ``Who are the crowdworkers?: shifting demographics in mechanical
  turk''.
\newblock In CHI'10 extended abstracts on Human factors in computing systems,
  ACM, pp.~2863--2872.

\bibitem{ostergaard2011does}
{\O}stergaard, C.~R., Timmermans, B., and Kristinsson, K., 2011.
\newblock ``Does a different view create something new? the effect of employee
  diversity on innovation''.
\newblock {\em Research Policy, \textbf{ 40}}(3), pp.~500--509.

\bibitem{horwitz2007effects}
Horwitz, S.~K., and Horwitz, I.~B., 2007.
\newblock ``The effects of team diversity on team outcomes: A meta-analytic
  review of team demography''.
\newblock {\em Journal of Management, \textbf{ 33}}(6), pp.~987--1015.

\bibitem{janis1971groupthink}
Janis, I.~L., 1971.
\newblock ``Groupthink''.
\newblock {\em Psychology today, \textbf{ 5}}(6), pp.~43--46.

\bibitem{williams1998demography}
Williams, K.~Y., and O'Reilly~III, C.~A., 1998.
\newblock ``Demography and''.
\newblock {\em Research in Organizational Behavior, \textbf{ 20}}, pp.~77--140.

\bibitem{harrison2002time}
Harrison, D.~A., Price, K.~H., Gavin, J.~H., and Florey, A.~T., 2002.
\newblock ``Time, teams, and task performance: Changing effects of surface-and
  deep-level diversity on group functioning''.
\newblock {\em Academy of Management Journal, \textbf{ 45}}(5), pp.~1029--1045.

\bibitem{pelled1999exploring}
Pelled, L.~H., Eisenhardt, K.~M., and Xin, K.~R., 1999.
\newblock ``Exploring the black box: An analysis of work group diversity,
  conflict and performance''.
\newblock {\em Administrative Science Quarterly, \textbf{ 44}}(1), pp.~1--28.

\bibitem{cox1991managing}
Cox, T.~H., and Blake, S., 1991.
\newblock ``Managing cultural diversity: Implications for organizational
  competitiveness''.
\newblock {\em The Executive}, pp.~45--56.

\bibitem{bryne1966effect}
Bryne, D., Clore~Jr, G., and Worchel, P., 1966.
\newblock ``The effect of economic similarity-dissimilarity as determinants of
  attraction''.
\newblock {\em Journal of Personality and Social Psychology, \textbf{ 4}},
  pp.~220--224.

\bibitem{Lin11:Class}
Lin, H., and Bilmes, J., 2011.
\newblock ``A class of submodular functions for document summarization''.
\newblock In Annual Meeting of the Association for Computational Linguistics
  (ACL-HLT).

\bibitem{lin2012learning}
Lin, H., and Bilmes, J., 2012.
\newblock ``Learning mixtures of submodular shells with application to document
  summarization''.
\newblock In Proceedings of the Twenty-Eighth Conference on Uncertainty in
  Artificial Intelligence, pp.~479--490.

\bibitem{carbonell1998use}
Carbonell, J., and Goldstein, J., 1998.
\newblock ``The use of {MMR}, diversity-based reranking for reordering
  documents and producing summaries''.
\newblock In Proceedings of the 21st Annual International ACM SIGIR Conference
  on Research and Development in Information Retrieval, ACM, pp.~335--336.

\bibitem{zhu2007improving}
Zhu, X., Goldberg, A.~B., Van~Gael, J., and Andrzejewski, D., 2007.
\newblock ``Improving diversity in ranking using absorbing random walks.''.
\newblock In HLT-NAACL, Citeseer, pp.~97--104.

\bibitem{zhai2003beyond}
Zhai, C.~X., Cohen, W.~W., and Lafferty, J., 2003.
\newblock ``Beyond independent relevance: methods and evaluation metrics for
  subtopic retrieval''.
\newblock In Proceedings of the 26th Annual International ACM SIGIR Conference
  on Research and Development in Informaion Retrieval, ACM, pp.~10--17.

\bibitem{kulesza2012determinantal}
Kulesza, A., Taskar, B., et~al., 2012.
\newblock ``Determinantal point processes for machine learning''.
\newblock {\em Foundations and Trends{\textregistered} in Machine Learning,
  \textbf{ 5}}(2--3), pp.~123--286.

\bibitem{ahmed2018ranking}
Ahmed, F., and Fuge, M., 2018.
\newblock ``Ranking ideas for diversity and quality''.
\newblock {\em Journal of Mechanical Design, \textbf{ 140}}(1), p.~011101.

\bibitem{wilde2008teamology}
Wilde, D.~J., 2008.
\newblock {\em Teamology: The Construction and Organization of Effective
  Teams}.
\newblock Springer London.

\bibitem{cruz2014group}
Cruz, W.~M., and Isotani, S., 2014.
\newblock ``Group formation algorithms in collaborative learning contexts: A
  systematic mapping of the literature''.
\newblock In CYTED-RITOS International Workshop on Groupware, Springer,
  pp.~199--214.

\bibitem{anagnostopoulos2012online}
Anagnostopoulos, A., Becchetti, L., Castillo, C., Gionis, A., and Leonardi, S.,
  2012.
\newblock ``Online team formation in social networks''.
\newblock In Proceedings of the 21st international conference on World Wide
  Web, ACM, pp.~839--848.

\bibitem{ijcai2017-6}
Ahmed, F., Dickerson, J.~P., and Fuge, M., 2017.
\newblock ``Diverse weighted bipartite b-matching''.
\newblock In Proceedings of the International Joint Conference on Artificial
  Intelligence (IJCAI), pp.~35--41.

\bibitem{ahmadi2019algorithms}
Ahmadi, S., Ahmed, F., Dickerson, J.~P., Fuge, M., and Khuller, S., 2019.
\newblock ``An algorithm for multi-attribute diverse matching''.
\newblock {\em arXiv preprint arXiv:1909.03350}.

\bibitem{Cohen:2017:CDG:3068839.3068842}
Cohen, S., and Yashinski, M., 2017.
\newblock ``Crowdsourcing with diverse groups of users''.
\newblock In Proceedings of the 20th International Workshop on the Web and
  Databases, WebDB'17, ACM, pp.~7--12.

\bibitem{basu2015task}
Basu~Roy, S., Lykourentzou, I., Thirumuruganathan, S., Amer-Yahia, S., and Das,
  G., 2015.
\newblock ``Task assignment optimization in knowledge-intensive
  crowdsourcing''.
\newblock {\em The VLDB Journal—The International Journal on Very Large Data
  Bases, \textbf{ 24}}(4), pp.~467--491.

\bibitem{schmitz2018online}
Schmitz, H., and Lykourentzou, I., 2018.
\newblock ``Online sequencing of non-decomposable macrotasks in expert
  crowdsourcing''.
\newblock {\em ACM Transactions on Social Computing, \textbf{ 1}}(1), p.~1.

\bibitem{Karp90:Optimal}
Karp, R.~M., Vazirani, U.~V., and Vazirani, V.~V., 1990.
\newblock ``An optimal algorithm for on-line bipartite matching''.
\newblock In Proceedings of the Annual Symposium on Theory of Computing (STOC),
  pp.~352--358.

\bibitem{mirzasoleiman2018streaming}
Mirzasoleiman, B., Jegelka, S.~S., and Krause, A., 2018.
\newblock ``Streaming non-monotone submodular maximization: Personalized video
  summarization on the fly''.
\newblock In AAAI Conference on Artificial Intelligence 2018, Association for
  the Advancement of Artificial Intelligence (AAAI).

\bibitem{Devanur12:Online}
Devanur, N.~R., and Jain, K., 2012.
\newblock ``Online matching with concave returns''.
\newblock In Proceedings of ACM Symposium on Theory of Computing (STOC), ACM,
  pp.~137--144.

\bibitem{Agrawal14:Fast}
Agrawal, S., and Devanur, N.~R., 2014.
\newblock ``Fast algorithms for online stochastic convex programming''.
\newblock In Proceedings of the ACM-SIAM Symposium on Discrete Algorithms
  (SODA), SIAM, pp.~1405--1424.

\bibitem{7906050}
Yu, Q., Xu, E.~L., and Cui, S., 2016.
\newblock ``Submodular maximization with multi-knapsack constraints and its
  applications in scientific literature recommendations''.
\newblock In 2016 IEEE Global Conference on Signal and Information Processing
  (GlobalSIP), pp.~1295--1299.

\bibitem{badanidiyuru2014fast}
Badanidiyuru, A., and Vondr{\'a}k, J., 2014.
\newblock ``Fast algorithms for maximizing submodular functions''.
\newblock In Annual ACM-SIAM Symposium on Discrete Algorithms (SODA),
  pp.~1497--1514.

\bibitem{ahmed2019measuring}
Ahmed, F., Ramachandran, S.~K., Fuge, M., Hunter, S., and Miller, S., 2019.
\newblock ``Measuring and optimizing design variety using herfindahl index''.
\newblock In ASME 2019 International Design Engineering Technical Conferences
  and Computers and Information in Engineering Conference, American Society of
  Mechanical Engineers Digital Collection.

\bibitem{yu2016streaming}
Yu, Q., Xu, L., and Cui, S., 2018.
\newblock ``Streaming algorithms for news and scientific literature
  recommendation: Monotone submodular maximization with a $ d $-knapsack
  constraint''.
\newblock {\em IEEE Access, \textbf{ 6}}, pp.~53736--53747.

\bibitem{lubin2016polyhedral}
Lubin, M., Yamangil, E., Bent, R., and Vielma, J.~P., 2016.
\newblock ``Polyhedral approximation in mixed-integer convex optimization''.
\newblock {\em Mathematical Programming}, pp.~1--30.

\bibitem{DiNoia:2014:AUP:2645710.2645774}
Di~Noia, T., Ostuni, V.~C., Rosati, J., Tomeo, P., and Di~Sciascio, E., 2014.
\newblock ``An analysis of users' propensity toward diversity in
  recommendations''.
\newblock In ACM Conference on Recommender Systems (RecSys), pp.~285--288.

\bibitem{Difallah:2018:DDM:3159652.3159661}
Difallah, D., Filatova, E., and Ipeirotis, P., 2018.
\newblock ``Demographics and dynamics of mechanical turk workers''.
\newblock In Proceedings of the ACM International Conference on Web Search and
  Data Mining (WSDM), ACM, pp.~135--143.

\bibitem{doi:10.1177/2158244016636433}
Levay, K.~E., Freese, J., and Druckman, J.~N., 2016.
\newblock ``The demographic and political composition of mechanical turk
  samples''.
\newblock {\em SAGE Open, \textbf{ 6}}(1), p.~2158244016636433.

\bibitem{siangliulue2015toward}
Siangliulue, P., Arnold, K.~C., Gajos, K.~Z., and Dow, S.~P., 2015.
\newblock ``Toward collaborative ideation at scale: Leveraging ideas from
  others to generate more creative and diverse ideas''.
\newblock In Proceedings of the 18th ACM Conference on Computer Supported
  Cooperative Work \& Social Computing, ACM, pp.~937--945.

\bibitem{kim2014ensemble}
Kim, J., Cheng, J., and Bernstein, M.~S., 2014.
\newblock ``Ensemble: exploring complementary strengths of leaders and crowds
  in creative collaboration''.
\newblock In Proceedings of the 17th ACM conference on Computer supported
  cooperative work \& social computing, ACM, pp.~745--755.

\bibitem{ye2017does}
Ye, T., and Robert~Jr, L.~P., 2017.
\newblock ``Does collectivism inhibit individual creativity?: The effects of
  collectivism and perceived diversity on individual creativity and
  satisfaction in virtual ideation teams''.
\newblock In Proceedings of the 2017 ACM Conference on Computer Supported
  Cooperative Work and Social Computing, ACM, pp.~2344--2358.

\bibitem{dong2016embracing}
Dong, W., Ehrlich, K., Macy, M.~M., and Muller, M., 2016.
\newblock ``Embracing cultural diversity: Online social ties in distributed
  workgroups''.
\newblock In Proceedings of the 19th ACM Conference on Computer-Supported
  Cooperative Work \& Social Computing, ACM, pp.~274--287.

\bibitem{allahbakhsh2013quality}
Allahbakhsh, M., Benatallah, B., Ignjatovic, A., Motahari-Nezhad, H.~R.,
  Bertino, E., and Dustdar, S., 2013.
\newblock ``Quality control in crowdsourcing systems: Issues and directions''.
\newblock {\em IEEE Internet Computing, \textbf{ 17}}(2), pp.~76--81.

\bibitem{fuge:openideo_collaboration}
Fuge, M., Tee, K., Agogino, A., and Maton, N., 2014.
\newblock ``Analysis of collaborative design networks: A case study of
  openideo''.
\newblock {\em Journal of Computing and Information Science in Engineering,
  \textbf{ 14}}(2), p.~021009.

\bibitem{fuge:idetc_2014_oi}
Fuge, M., and Agogino, A., 2014.
\newblock ``How online design communities evolve over time: the birth and
  growth of {OpenIDEO}''.
\newblock In ASME International Design Engineering Technical Conferences, ASME.

\bibitem{moreno2015bayesian}
Moreno, P.~G., Art{\'e}s-Rodr{\'\i}guez, A., Teh, Y.~W., and Perez-Cruz, F.,
  2015.
\newblock ``Bayesian nonparametric crowdsourcing''.
\newblock {\em The Journal of Machine Learning Research, \textbf{ 16}}(1),
  pp.~1607--1627.

\end{thebibliography}


\begin{table*}[htbp]
\caption{Table of notation}
\begin{center}
\small
\begin{tabular}{r p{0.1cm} p{10cm} p{1.6cm} }
\toprule
Parameters &  & Description & Values in Sec. 5.1\\
\midrule
$G$  & $\triangleq$ & Bipartite graph & \\
$P$  & $\triangleq$ & Set of vertices that arrive sequentially (\eg ~people) &\\
$T$ & $\triangleq$ & Set of all vertices known apriori (\eg ~teams) &  \\
$E$ & $\triangleq$ & Set of all edges &\\
$S$  & $\triangleq$ &   A feasible team allocation & \\
$y_{k,j}$ & $\triangleq$ &   Number of people from cluster $k$, allocated to team $j$ & \\
$f(S_j)$ & $\triangleq$ &  Objective function value for a set of people matched to a team $j$ & \\
$ \Delta f(S, e)$ & $\triangleq$ &  Marginal gain on adding edge e to set S, which is $f(S \cup e) - f(S)$& \\
$S_j$ & $\triangleq$ & Subset of edges in a matching that are incident to vertex $j$ & \\
$P_k$  & $\triangleq$ & Set of people that belong to cluster k & \\
$N$  & $\triangleq$ & Number of teams to be formed & 10 \\
$M$  & $\triangleq$ & Maximum number of people that can arrive sequentially & 100\\
$K$  & $\triangleq$ &  Number of clusters into which nodes arriving sequentially are partitioned & 3 \\
$b(i)$  & $\triangleq$ & Maximum number of edges that can be matched to each node (team or people) & 3\\
$R^+$  & $\triangleq$ & Maximum people that can be matched to each team & 3\\
$L^+$  & $\triangleq$ & Maximum teams that can be matched to each person & N/A \\
$w_{i,j}$ & $\triangleq$ & Utility of person $i$ for team $j$ & 1.0 \\
$d$  & $\triangleq$ &   Total number of knapsack constraints & 10 \\
$c_{e, j j}$ & $\triangleq$ & Cardinality cost of an edge  $e$ (from person $i$ to team $j$) for each constraint & 1 \\
$df_{R^-}$ & $\triangleq$ &  $R^-$th highest marginal gain among all clusters & 1.0 \\
$c_{i,j}^B$  & $\triangleq$ &   Payment to a person i for team j after acceptance &  N/A\\
$c_i^S$  & $\triangleq$ &   Cost of interviewing a person i &  N/A\\
$B$  & $\triangleq$ &   Total budget of a firm hiring people & N/A\\
$OPT^{*}$ & $\triangleq$ &   Optimal utility for offline matching & 30.0\\
$\alpha$ & $\triangleq$ &  An algorithm parameter between 0 to 1, which determines the marginal gain cutoff & 1.0\\
$v$ & $\triangleq$ &  A value selected between $\alpha OPT*$ and $OPT*$ & 30.0\\
\bottomrule
\end{tabular}
\end{center}
\label{tab:TableOfNotation3}
\end{table*}
\end{document}